\documentclass[10pt,twocolumn,letterpaper]{article}
\PassOptionsToPackage{table}{xcolor}
\usepackage[pagenumbers]{cvpr}
\usepackage{times}
\def\confName{CVPR}
\def\confYear{2027}

\usepackage[utf8]{inputenc} 
\usepackage[T1]{fontenc}    
\usepackage{courier}        
\usepackage{url}            
\usepackage{booktabs}       
\usepackage{amsmath}
\usepackage{amsfonts}       
\usepackage{nicefrac}       
\usepackage{microtype}      
\usepackage{xcolor}         
\makeatletter
\def\@fnsymbol#1{\ensuremath{\ifcase#1\or \dagger\or \ddagger\or
   \mathsection\or \mathparagraph\or \|\or **\or \dagger\dagger
   \or \ddagger\ddagger \else\@ctrerr\fi}}
\makeatother

\usepackage{multirow}
\usepackage{algorithm}
\usepackage{algorithmic}
\usepackage[most]{tcolorbox}
\tcbuselibrary{skins,breakable}
\usepackage{graphicx}
\usepackage{float}
\usepackage{tikz}
\usepackage{amsthm}

\definecolor{cvprblue}{rgb}{0.21,0.49,0.74}
\definecolor{Red}{RGB}{255, 199, 206}
\definecolor{Blue}{RGB}{221, 235, 247}
\definecolor{Yellow}{RGB}{255, 235, 156}
\definecolor{softgreen}{RGB}{34,139,34}

\usepackage[pagebackref,breaklinks,colorlinks,allcolors=cvprblue]{hyperref}
\hypersetup{
  pdftitle={DecFlowEdit: Self-Localized Flow-based Image Editing via Guidance Decoupling},
  pdfauthor={Zheyuan Zhan, Can Wang, Jiawei Chen, Chun Chen, Siwei Lyu, Zeyu Zheng, Defang Chen}
}
\let\cite\citep

\title{
DecFlowEdit: Self-Localized Flow-based Image Editing via Guidance Decoupling
}
\author{\textbf{Zheyuan Zhan}$^{1}$\quad
  \textbf{Can Wang}$^{1}$\quad
  \textbf{Jiawei Chen}$^{1}$\quad
  \textbf{Chun Chen}$^{1}$\quad
  \textbf{Siwei Lyu}$^{2}$\\[0.25em]
  \textbf{Zeyu Zheng}$^{3}$\qquad
  \textbf{Defang Chen}$^{3}$\thanks{Corresponding author.}\\[0.5em]
  $^{1}$\normalfont{Zhejiang University}\quad
  $^{2}$\normalfont{University at Buffalo}\quad
  $^{3}$\normalfont{University of California, Berkeley}
}

\begin{document}

\maketitle
\suppressfloats[t] 

\begin{abstract}

Flow-based image editing (FlowEdit) enables inversion-free semantic changes through the difference between source and target velocities. In this paper, we observe that FlowEdit's default classifier-free guidance (CFG) configuration, with asymmetric source and target scales, causes substantial background leakage. Matching these guidance scales, for example by removing CFG, improves edit-relevant localization but severely degrades editability. To get the best of both worlds, we propose DecFlowEdit, which decouples the optimal guidance scales for localization and for editing in flow-based generative models. In particular, DecFlowEdit first extracts an edit-relevant prior by temporally aggregating velocity differences evaluated without CFG, and then uses this prior to reweight the original updates under default CFG. Our method remains training-free and inversion-free, requiring neither external spatial masks nor attention manipulation. Experiments on PIE-Bench across FLUX.1-dev, SD3, and SD3.5 show that DecFlowEdit improves background preservation, reducing structure distance by approximately 61--73\% and background LPIPS by 68--80\% relative to FlowEdit at comparable editing fidelity.

\end{abstract}

\section{Introduction} \label{sec:intro}

Text-guided image editing aims to realize a user-specified semantic change while preserving edit-irrelevant content. Achieving both objectives requires control over where the changes occur. Training-based methods learn this behavior from editing data~\cite{brook2023instructpix2pix,geng2023instructdiffusion}, whereas training-free methods exploit pretrained generative models without task-specific optimization. Despite their flexibility, training-free methods often face a trade-off: stronger semantic changes can introduce unintended modifications outside the edit region.

\begin{figure*}[t]
    \centering
    \includegraphics[width=1.00\linewidth]{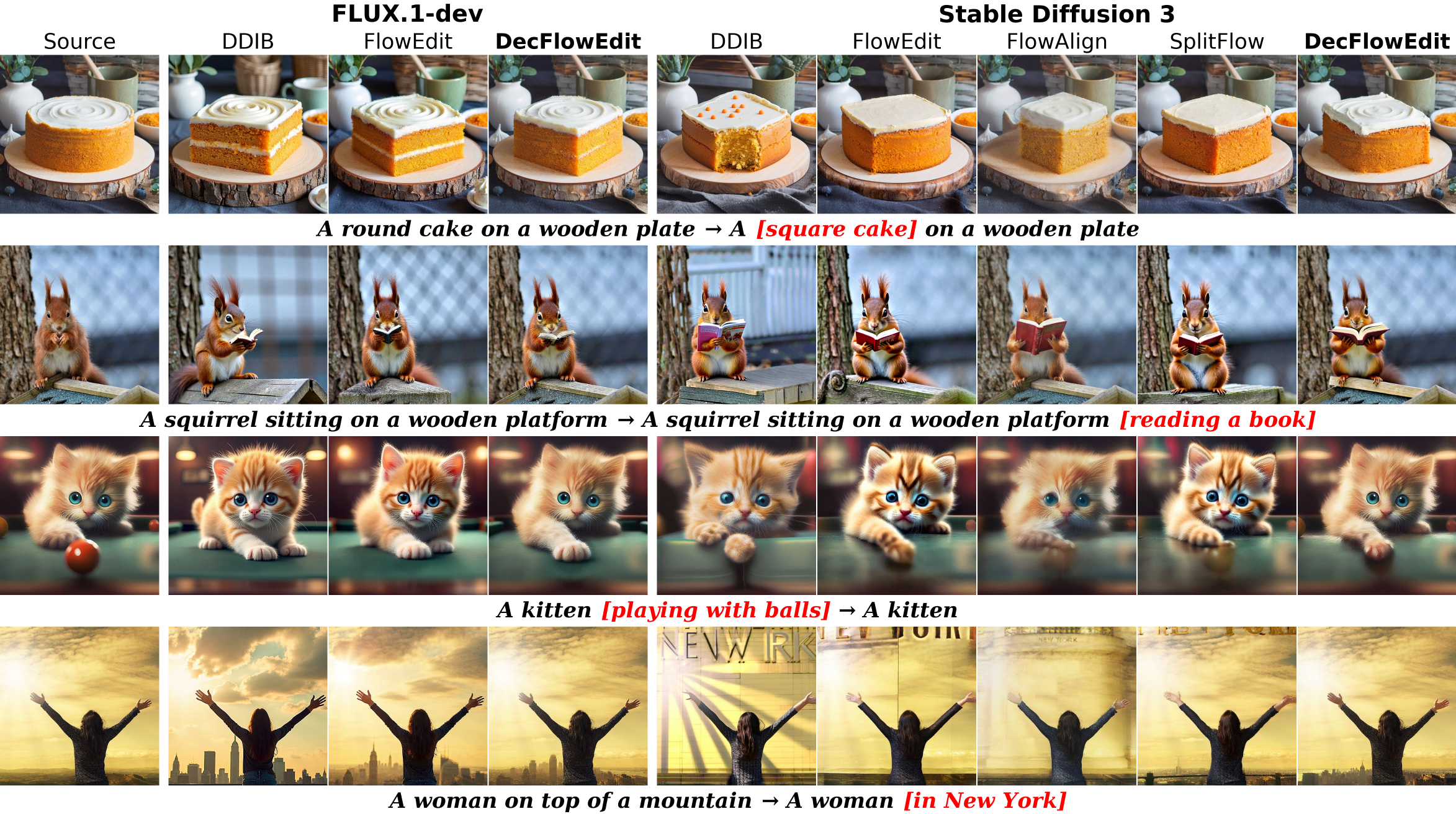}
    \caption{Qualitative comparison on PIE-Bench using FLUX.1-dev (left) and SD3 (right). Text below each row shows source-to-target prompt changes, with edited phrases in red brackets. DecFlowEdit edits as intended while better preserving non-edited content and structure.}
    \label{fig:qualitative}
\end{figure*}

Many training-free methods follow an inversion-and-regeneration paradigm, first mapping the source image to a latent or noise representation and then regenerating it under the target condition~\cite{su2023ddib}. However, preserving source content remains challenging, since these methods regenerate the whole image, including regions that should remain unchanged. Attention- and feature-control methods, including Prompt-to-Prompt~\cite{hertz2023p2p} and subsequent approaches~\cite{tumanyan2023plug,cao2023masactrl}, improve background preservation by injecting source information into the generation process. Such interventions require access to internal model architectures and increase inference complexity. External spatial localization can also constrain the edit region, but it usually relies on additional localization signals~\cite{couairon2023diffedit,zhu2025kvedit} or external segmentation models~\cite{li2023layerdiffusion}, which may be unavailable or may not align with the regions affected by the editing dynamics. Built on flow matching~\cite{lipman2023flow}, flow-based image editing (FlowEdit)~\cite{kulikov2025flowedit} constructs paired stochastic source and target states and integrates their conditional velocity differences into a direct editing trajectory, without inversion or attention manipulation. Compared with inversion-and-regeneration~\cite{su2023ddib}, this formulation improves source preservation, but its updates remain spatially global. Although its velocity differences are typically stronger in edit-relevant regions, residual differences in non-edit regions may accumulate across timesteps, causing background leakage (see Figure~\ref{fig:qualitative}).

\begin{figure*}[!t]
    \centering
    \includegraphics[width=\linewidth]
    {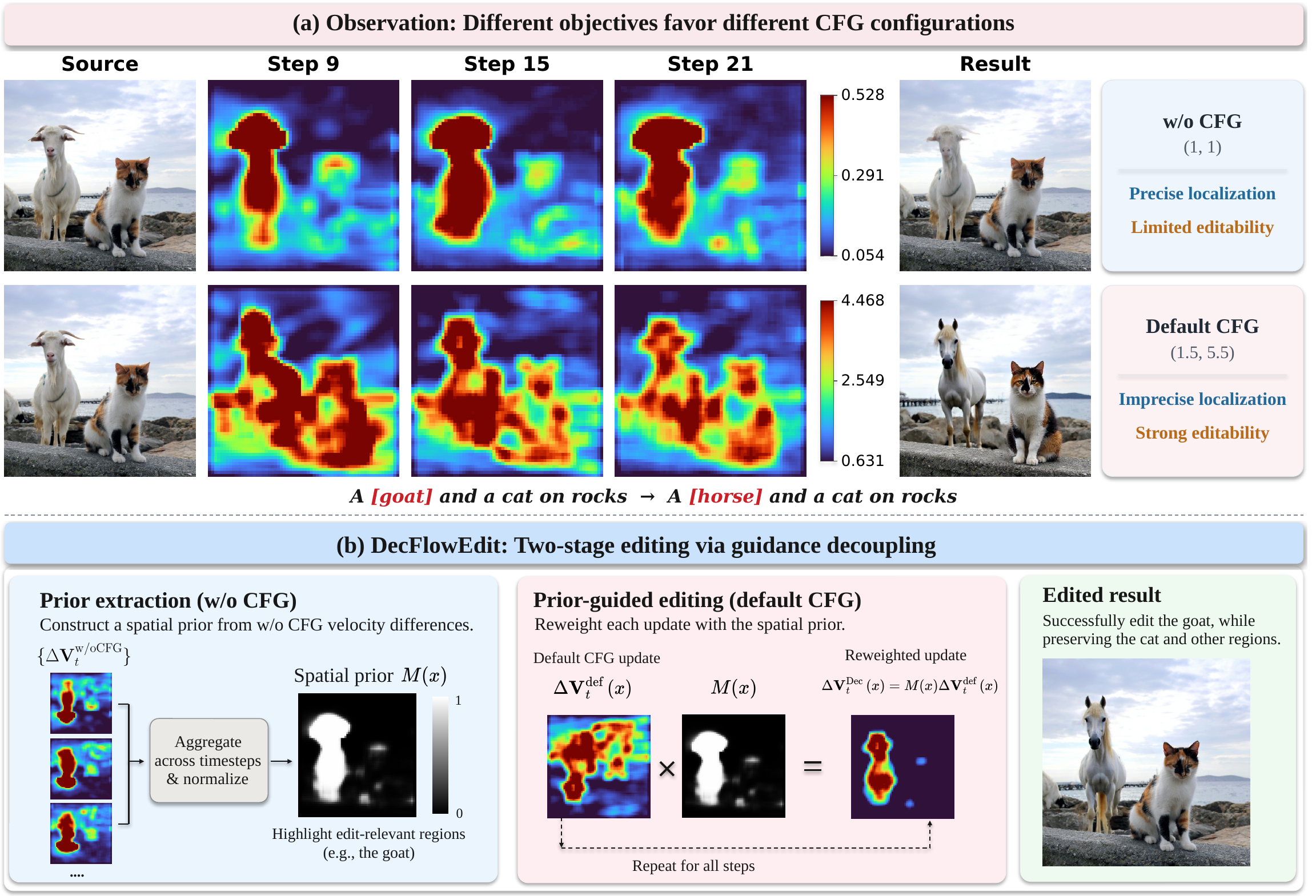}
    \caption{
    \textbf{Observation and overview of DecFlowEdit (FLUX.1-dev).}
    (a) Velocity-difference magnitude maps and editing results under w/o~CFG and default CFG.
    Colorbars indicate magnitude scales.
    (b) Prior extraction under w/o~CFG and prior-guided editing under default CFG.
    }
    \label{fig:guidance_observation}
\end{figure*}

In this work, we identify the source and target classifier-free guidance (CFG) scales as a key factor governing this leakage. By default, FlowEdit uses asymmetric source and target guidance scales (default CFG) to achieve effective semantic editing. However, this asymmetry creates mismatched source and target velocity responses even in semantically unchanged regions, weakening their cancellation in the velocity difference. We observe that matching the guidance scales improves edit-relevant localization, especially when both scales are set to one, which removes CFG amplification. Without CFG (w/o~CFG), velocity differences concentrate more closely on the intended edit region, but directly integrating them produces weak semantic changes (see Figure~\ref{fig:guidance_observation}(a)). Thus, the guidance configuration for localization differs from that for effective editing.

Motivated by this observation, we propose \textbf{DecFlowEdit}, which decouples guidance for spatial localization from guidance for semantic editing. DecFlowEdit first extracts an edit-relevant spatial prior by temporally aggregating source--target velocity differences evaluated under w/o~CFG. We then use the resulting soft prior to spatially reweight the FlowEdit updates under default CFG. This design separates the estimation of where to edit from the generation of the semantic change: the w/o~CFG responses supply spatial localization, while the default CFG updates retain the strength needed for editing. Our main contributions are as follows: 
\begin{itemize}
    \item We identify a guidance-dependent trade-off between localization and editing in flow-based image editing: default CFG produces effective semantic changes but substantial background leakage, whereas w/o~CFG improves localization at the cost of editability. 
    \item We develop DecFlowEdit, which extracts an intrinsic spatial prior by temporally aggregating velocity differences under w/o~CFG and uses it to reweight editing updates under default CFG. The method requires no training, inversion, external masks, or attention manipulation. 
    \item We demonstrate consistent improvements in background and structure preservation on PIE-Bench across FLUX.1-dev, SD3, and SD3.5 with comparable editing fidelity, supporting guidance decoupling for effective spatial control.
\end{itemize}
\section{Related Work} \label{sec:related_works}


\paragraph{Inversion and attention manipulation.}
A line of training-free real-image editing methods follows the inversion-and-regeneration paradigm of Dual Diffusion Implicit Bridges (DDIB), where the source image is first mapped to a latent noise representation and then regenerated under the target condition~\cite{su2023ddib}. Within this paradigm, Prompt-to-Prompt~\cite{hertz2023p2p} introduces cross-attention replacement to preserve spatial layout, inspiring subsequent attention- and feature-manipulation methods~\cite{tumanyan2023plug,cao2023masactrl,wang2025taming,jiao2026unieditflow}. While effective in improving structure preservation, these methods have three practical drawbacks. First, their behavior is sensitive to the choice of intervention layers and timesteps; inappropriate schedules or large source--target semantic gaps can introduce artifacts when preserved source patterns conflict with the intended target semantics. Second, they require overriding internal attention computations and maintaining source-branch representations during sampling, increasing implementation complexity and potentially incurring substantial inference overhead. Third, adapting these interventions to different backbones typically requires architecture-specific redesign.


\paragraph{FlowEdit and preservation-oriented flow editing.}
Built on flow matching models, FlowEdit constructs an inversion-free editing path from the difference between target- and source-conditioned velocity fields, yielding a more direct path between the source and target images and improved background preservation compared with inversion-based editing~\cite{su2023ddib,lipman2023flow,esser2024scaling,black-forest2024flux,kulikov2025flowedit}. Recent follow-ups further improve content preservation and editing consistency through trajectory regularization~\cite{kim2026flowalign}, direct noise alignment~\cite{xie2025dnaedit}, target-aware intermediate states~\cite{wang2025flowcycle}, or semantic flow decomposition and velocity correction~\cite{yoon2025splitflow,li2026cvc,endo2026flowslider}. CVC~\cite{li2026cvc} and FlowSlider~\cite{endo2026flowslider} separate structural and semantic components within an editing update, whereas DecFlowEdit decouples localization and editing across CFG configurations. We observe that w/o~CFG produces more localized velocity differences but weak semantic edits, while default CFG enables effective editing but can cause background leakage. Accordingly, we extract a spatial prior under w/o~CFG and use it to constrain updates \mbox{under default CFG}.

\paragraph{Spatial localization for background preservation.}
Spatial localization provides explicit control over the regions affected by image editing. Attention-based methods extract regions from internal attention maps~\cite{cao2023masactrl,patashnik2023localizing}, requiring access to the model's internal attention representations. Other approaches rely on user-provided masks or external segmentation models~\cite{kirillov2023segment,li2023layerdiffusion,zhu2025kvedit}, introducing manual input or an additional model dependency. Prediction-based methods infer edit regions from source--target differences in noise predictions, as in DiffEdit~\cite{couairon2023diffedit}, or in velocity predictions, as in UniEdit-Flow~\cite{jiao2026unieditflow}; because the contrast is taken on high-variance per-step predictions, the resulting masks are unstable and leak into non-edit regions. In contrast, our method obtains localization from FlowEdit's own velocity differences, which are more spatially localized w/o~CFG. We aggregate these responses along a w/o~CFG trajectory into a soft spatial prior. This temporal aggregation makes the prior more stable than one extracted from a single timestep. The resulting prior provides spatial control over editing without requiring attention manipulation, external segmentation models, or user-provided masks.

\section{Background}
\label{sec:background}

\paragraph{Conditional flow-based generation.}
Recent training-free editing methods are increasingly built upon flow matching models, which learn a continuous mapping from a data distribution $\mathbf{x}_0 \sim p_0$ to a prior $\mathbf{x}_1 \sim p_1$, typically the standard Gaussian $\mathcal{N}(\mathbf{0},\mathbf{I})$~\cite{lipman2023flow}. A linear interpolation
$\mathbf{x}_{t} = (1-t)\mathbf{x}_0 + t\mathbf{x}_1$, $t \in [0,1]$, defines the forward trajectory, and the trajectory follows $d\mathbf{x}_t = \mathbf{V}_{\theta}(\mathbf{x}_t, t \mid P)\,dt$, where $\mathbf{V}_{\theta}$ is the conditional velocity field and $P$ denotes the textual condition. Flow matching trains the time-dependent conditional velocity field by minimizing
$\mathcal{L}_{\theta}
=
\mathbb{E}_{t,\mathbf{x}_0,\mathbf{x}_1,P}
[
\left\|
(\mathbf{x}_1-\mathbf{x}_0)
-
\mathbf{V}_{\theta}(\mathbf{x}_t,t\mid P)
\right\|^2
]$.

\paragraph{Inversion-based editing using flow-based models.}
DDIB-style inversion-and-regeneration editing maps a source image to a noise latent under the source prompt and then regenerates it under the target prompt~\cite{su2023ddib}. Let $P_{\mathrm{src}}$ and $P_{\mathrm{tar}}$ denote the source and target prompts. The source-side inversion trajectory is obtained by integrating
$d\mathbf{z}^{\mathrm{src}}_t
=
\mathbf{V}_{\theta}(\mathbf{z}^{\mathrm{src}}_t,t\mid P_{\mathrm{src}})\,dt$,
starting from the source image $\mathbf{z}^{\mathrm{src}}_0:=\mathbf{x}^{\mathrm{src}}$ and proceeding toward a noise latent $\mathbf{z}^{\mathrm{src}}_1$. Editing then integrates
$d\mathbf{z}^{\mathrm{tar}}_t
=
\mathbf{V}_{\theta}(\mathbf{z}^{\mathrm{tar}}_t,t\mid P_{\mathrm{tar}})\,dt$
backward from the shared noise latent $\mathbf{z}^{\mathrm{tar}}_1=\mathbf{z}^{\mathrm{src}}_1$ to obtain the edited image $\mathbf{z}^{\mathrm{tar}}_0$. This shared latent provides only coarse structural control, and target-conditioned regeneration often introduces substantial changes to details.

\paragraph{Inversion-free editing using velocity differences.}
FlowEdit constructs a direct editing path by pairing source and target states at each timestep, without mapping the source image to Gaussian noise~\cite{kulikov2025flowedit}. Following FlowEdit's time convention, the editing path starts from the source image $\mathbf{z}^{\mathrm{edit}}_1:=\mathbf{x}^{\mathrm{src}}$ and is integrated backward toward the edited image $\mathbf{z}^{\mathrm{edit}}_0$. At timestep $t$, given the current editing state $\mathbf{z}^{\mathrm{edit}}_t$ and freshly sampled Gaussian noise $\boldsymbol{\epsilon}_t\sim\mathcal{N}(\mathbf{0},\mathbf{I})$, the paired noisy states are
$\mathbf{z}^{\mathrm{src}}_t
=
(1-t)\mathbf{x}^{\mathrm{src}}+t\boldsymbol{\epsilon}_t,
\qquad
\mathbf{z}^{\mathrm{tar}}_t
=
\mathbf{z}^{\mathrm{edit}}_t
+
\mathbf{z}^{\mathrm{src}}_t
-
\mathbf{x}^{\mathrm{src}}.$
FlowEdit uses the difference between the two conditional velocity fields as the editing direction,
$\Delta\mathbf{V}_t
=
\mathbf{V}_{\theta}
(\mathbf{z}^{\mathrm{tar}}_t,t\mid P_{\mathrm{tar}})
-
\mathbf{V}_{\theta}
(\mathbf{z}^{\mathrm{src}}_t,t\mid P_{\mathrm{src}}),$
and updates the editing trajectory by integrating
$d\mathbf{z}^{\mathrm{edit}}_t=\Delta\mathbf{V}_t\,dt$.
Sharing the same noise within each timestep promotes similar source and target velocity responses in edit-irrelevant regions, facilitating their cancellation in the velocity difference. This construction keeps editing anchored to the source image and improves content preservation compared with inversion-and-regeneration. In practice, FlowEdit relies on strong, asymmetric source and target CFG scales for effective semantic editing. Asymmetric guidance can introduce a mismatch between the source and target responses even in semantically unchanged regions, weakening their cancellation. Stronger guidance can further amplify residual updates in these regions. Consequently, effective semantic editing can be accompanied by background leakage.
\section{Method}
\label{sec:method}

We first examine how CFG affects spatial localization and semantic
editing in FlowEdit (Section~\ref{subsec:guidance_observation}). We then
describe the two stages of DecFlowEdit: \emph{prior extraction}, which
constructs a soft spatial prior from velocity differences under w/o~CFG
(Section~\ref{subsec:without_cfg_prior}), and \emph{prior-guided editing},
which uses this prior to spatially constrain the updates under default CFG
(Section~\ref{subsec:default_cfg_editing}).

\subsection{Decoupling Localization from Editing}
\label{subsec:guidance_observation}

Let
$\mathbf{g}=(g_{\mathrm{src}},g_{\mathrm{tar}})$
denote the source and target CFG scales, and let
$\Delta\mathbf{V}^{\mathbf{g}}_t$
be the corresponding FlowEdit velocity difference.
We refer to
$\mathbf{g}^{\mathrm{w/oCFG}}=(1,1)$,
which applies no additional guidance amplification, as \emph{w/o~CFG}. We use \emph{default CFG} to denote
FlowEdit's backbone-specific default source and target CFG scales.

For each CFG configuration $\mathbf g$, Figure~\ref{fig:guidance_observation}(a)
visualizes the velocity-difference magnitude
$r_t^{\mathbf g}(x)=\|\Delta\mathbf V_t^{\mathbf g}(x)\|_2$,
where $x$ indexes a location in the latent spatial grid and the norm is taken
over channels. The two CFG configurations show complementary behavior.
As indicated by the colorbars, FlowEdit under w/o~CFG produces small-magnitude
velocity differences that remain concentrated on the goat across steps 9, 15,
and 21, leaving the final image almost unchanged. FlowEdit under default CFG
produces velocity differences with much larger magnitudes and replaces the goat
with a horse, but the updates also spread to the neighboring cat and cause an
unintended change.

To analyze this trade-off, let $\mathbf V_\theta^P$ and $\mathbf V_\theta^u$
denote the conditional and unconditional velocities evaluated at the current
state and time $t$ of each branch, where $P$ is its text prompt and $u$ denotes
the unconditional prediction. We define the guidance direction as
$\mathbf d_t^P=\mathbf V_\theta^P-\mathbf V_\theta^u$, and denote its source and
target counterparts by $\mathbf d_t^{\mathrm{src}}$ and
$\mathbf d_t^{\mathrm{tar}}$. Under standard linear CFG, the velocity difference
at a fixed pair of source and target states can be decomposed as
\begin{equation}
\resizebox{\dimexpr\linewidth-2.5em\relax}{!}{$\displaystyle
\Delta\mathbf V_t^{\mathbf g}=\Delta\mathbf V_t^{\mathrm{w/oCFG}}+\underbrace{(g_{\mathrm{tar}}-g_{\mathrm{src}})\mathbf d_t^{\mathrm{src}}}_{\text{\small asymmetry term}}+\underbrace{(g_{\mathrm{tar}}-1)\left(\mathbf d_t^{\mathrm{tar}}-\mathbf d_t^{\mathrm{src}}\right)}_{\text{\small amplification term}}.
$}
\label{eq:cfg_decomposition}
\end{equation}
The derivation and empirical analysis are provided in Appendices~\ref{app:cfg_decomposition_derivation} and~\ref{app:cfg_decomposition_empirical}, respectively.

The source guidance direction $\mathbf d_t^{\mathrm{src}}$ is not restricted to
the intended edit region and can contain responses in regions that should be
preserved. When $g_{\mathrm{tar}}\neq g_{\mathrm{src}}$, the nonzero coefficient
$g_{\mathrm{tar}}-g_{\mathrm{src}}$ introduces these responses into
$\Delta\mathbf V_t^{\mathbf g}$ through the asymmetry term, potentially causing
unintended changes. Matching the source and target guidance scales removes
this contribution.

The amplification term scales the remaining difference between the two guidance
directions, $\mathbf d_t^{\mathrm{tar}}-\mathbf d_t^{\mathrm{src}}$, by
$g_{\mathrm{tar}}-1$. When $g_{\mathrm{tar}}>1$, increasing it amplifies these
differences, including residual responses in non-edit regions. Thus, setting
$g_{\mathrm{src}}=g_{\mathrm{tar}}=g$ removes the asymmetry term but retains
$(g-1)(\mathbf d_t^{\mathrm{tar}}-\mathbf d_t^{\mathrm{src}})$.
At $(1,1)$, both additional terms vanish.

These complementary behaviors motivate our two-stage editing framework
(Figure~\ref{fig:guidance_observation}(b)): the prior-extraction stage extracts
a spatial prior under w/o~CFG, and the prior-guided editing stage uses it to
reweight updates under default CFG. This combines localized updates with
effective semantic editing: the goat is replaced with a horse, while the cat
and other regions are preserved.

\subsection{Prior Extraction w/o~CFG}
\label{subsec:without_cfg_prior}

In the prior-extraction stage, we compute velocity differences
$\Delta\mathbf{V}^{\mathrm{w/oCFG}}_t(x)$ along an auxiliary FlowEdit
trajectory $\mathbf{z}^{\mathrm{prior}}_t$ under w/o~CFG. As Figure~\ref{fig:guidance_observation} shows, these responses already reveal the intended edit region at individual intermediate steps.
To obtain a stable spatial prior, we aggregate velocity differences from the middle of the prior-extraction schedule, where spatial
localization is relatively stable, before taking the norm:
\begin{equation}
E(x)
=
\left\|
\sum_{t\in\mathcal{T}_{\mathrm{mid}}}
\Delta\mathbf{V}^{\mathrm{w/oCFG}}_t(x)
\right\|_2,
\label{eq:without_cfg_energy}
\end{equation}
where, for the full prior-extraction schedule,
$\mathcal{T}_{\mathrm{mid}}=\{t_k:\lfloor0.25T\rfloor\leq k\leq\lceil0.75T\rceil\}$,
$T$ is the total number of steps in the prior-extraction schedule, and
$t_k$ is the sampling time at step index $k$, with $k$ starting from zero.
We refer to $E(x)$ as the energy map.
We then smooth the resulting spatial map using a $5\times5$ mean filter $S$, yielding $
A(x)=S(E)(x)$.
In practice, this stage can use a reduced sampling schedule to reduce
computational cost. On FLUX.1-dev, SD3, and SD3.5, a schedule with
approximately one quarter as many sampling steps yields comparable editing and
preservation performance (Section~\ref{subsec:main_results}).

We normalize the smoothed energy map using its 10th and 90th percentiles,
then apply a sigmoid transformation with threshold $\tau$ and softness $s$
to obtain the soft spatial prior $M(x)\in[0,1]$.
The detailed construction is provided in Appendix~\ref{app:implementation_details}.

\subsection{Prior-guided Editing with Default CFG}
\label{subsec:default_cfg_editing}

In the prior-guided editing stage, we retain the spatial prior $M$ and
initialize a separate trajectory $\mathbf{z}^{\mathrm{edit}}_t$ from the
source image and update it under default CFG. At each step, we construct
the paired source and target states from $\mathbf{z}^{\mathrm{edit}}_t$
as in Section~\ref{sec:background}, evaluate their velocity difference $\Delta\mathbf{V}^{\mathrm{def}}_t$ under default CFG, and apply
the spatial prior:
\begin{equation}
\Delta\mathbf{V}^{\mathrm{Dec}}_t(x)
=
M(x)\,
\Delta\mathbf{V}^{\mathrm{def}}_t(x).
\label{eq:decflowedit_update}
\end{equation}
The scalar weight at each spatial location is broadcast across channels.
Locations with large weights retain more of the default CFG update, while
small weights suppress updates outside the edit region.

For consecutive sampling times $t_{k+1}\leq t_k$, the discrete update is
\begin{equation}
\resizebox{\dimexpr\linewidth-2.5em\relax}{!}{$\displaystyle
\mathbf{z}^{\mathrm{edit}}_{t_{k+1}}=\mathbf{z}^{\mathrm{edit}}_{t_k}+(t_{k+1}-t_k)\,M\odot\Delta\mathbf{V}^{\mathrm{def}}_{t_k},\qquad\mathbf{z}^{\mathrm{edit}}_{t_0}=\mathbf{x}^{\mathrm{src}}.
$}
\label{eq:decflowedit_integration}
\end{equation}
The velocity difference is re-evaluated along the spatially constrained trajectory at every step. We retain FlowEdit's paired-state construction
and integration rule, inserting spatial weighting before each update.
The same soft spatial prior is used throughout the
prior-guided editing stage.
\section{Experiments}
\label{sec:experiments}

\subsection{Experimental Setup}
\label{subsec:experimental_setup}

\paragraph{Dataset and metrics.}
We evaluate our method on the Prompt-based Image Editing Benchmark (PIE-Bench)~\cite{ju2023direct}, a widely used benchmark for text-guided image editing. PIE-Bench contains 700 editing samples covering diverse editing types, and each sample provides a source image, source and target prompts, an editing instruction, and a ground-truth edit mask. Following the standard PIE-Bench evaluation protocol, we report (1) structure distance based on DINO features, where a lower value indicates better structure preservation, (2) background preservation based on PSNR, LPIPS, MSE, and SSIM on the non-edit region specified by the ground-truth mask, and (3) semantic editing fidelity based on CLIP similarity on the whole image and on the edited region.

\paragraph{Baselines.}
We compare our method with recent training-free flow-based image editing methods, including FlowCycle~\cite{wang2025flowcycle}, SplitFlow~\cite{yoon2025splitflow}, FlowAlign~\cite{kim2026flowalign}, UniEdit-Flow~\cite{jiao2026unieditflow}, iRFDS~\cite{yang2024text}, DRFS~\cite{Beaudouin_2026_CVPR}, DNAEdit~\cite{xie2025dnaedit}, and FTEdit~\cite{xu2025fteedit}. For a controlled comparison, we further compare our method with the corresponding FlowEdit~\cite{kulikov2025flowedit} baseline on FLUX.1-dev~\cite{black-forest2024flux}, Stable Diffusion 3 (SD3) and 3.5 (SD3.5)~\cite{esser2024scaling}.


\paragraph{Implementation details.}
In the prior-guided editing stage, we follow FlowEdit's default CFG, using 28 sampling steps with source/target guidance scales of $(1.5,5.5)$ for FLUX.1-dev, and 50 sampling steps with $(3.5,13.5)$ for SD3/SD3.5. The prior-extraction stage uses FlowEdit under w/o~CFG, i.e., source/target scales of $(1,1)$. We construct the spatial prior as a soft map with threshold $\tau=0.55$ and softness parameter $s=0.08$. The main comparison uses the full sampling schedule for prior extraction (the same number of sampling steps as in the prior-guided editing stage). We apply the aggregation window only to executed steps. Following FlowEdit's default sampling settings, the full prior-extraction schedules execute steps 4--27 for FLUX.1-dev and 17--49 for SD3/SD3.5, using zero-based indexing. Applying the aggregation window yields indices 7--21 and 17--38, respectively, with both endpoints included. We also report a variant using a prior-extraction schedule with approximately one quarter as many sampling steps, while keeping the prior-guided editing stage fixed. See Appendix~\ref{app:implementation_details} for details.

\subsection{Main Results on PIE-Bench}
\label{subsec:main_results}

Table~\ref{tab:main_results} shows that DecFlowEdit yields consistent preservation gains: compared with raw FlowEdit, our method improves all background-preservation metrics and reduces structure distance by approximately 61--73\% across FLUX.1-dev, SD3, and SD3.5. These results show that the spatial prior extracted from w/o~CFG responses can constrain unintended changes in the final edit. The preservation gains are accompanied by a 2--3\% reduction in CLIP similarity. This reduction does not by itself indicate weaker edits: at comparable edited-region CLIP, spatial reweighting attains much lower background LPIPS than uniform update scaling (Appendix~\ref{app:spatial_uniform_control}), and constraining FlowEdit with ground-truth masks also lowers edited-region CLIP from 22.71 to 22.41 (Appendix~\ref{app:gt_mask_tradeoff}).
Across FLUX.1-dev, SD3, and SD3.5, DecFlowEdit retains strong content preservation and competitive text alignment with approximately one quarter as many prior-extraction sampling steps (Table~\ref{tab:main_results}). This shorter schedule reduces extraction time while retaining useful spatial localization. Relative to FlowEdit, the full method uses $2.0\times$ as many function evaluations (NFEs) on FLUX.1-dev and $1.5\times$ on SD3/SD3.5; Ours-short reduces these ratios to $1.25\times$ and $1.14\times$, respectively. Appendix~\ref{app:implementation_details} reports NFEs and measured end-to-end runtimes.

Figure~\ref{fig:qualitative} illustrates the effect of this spatial control. DecFlowEdit performs the intended semantic edits while better preserving background textures and non-edited objects than raw FlowEdit and other baselines. These results illustrate our two-stage design: the prior-extraction stage identifies where to edit, and the prior-guided editing stage performs the semantic edit under this spatial constraint.
We show how threshold, softness, and update strength affect the editing--preservation trade-off in Appendix~\ref{app:parameter_sensitivity} (Figures~\ref{fig:edit_preservation_tradeoff}--\ref{fig:parameter_s_scan}).

\begin{table*}[!t]
\centering
\caption{Quantitative comparison on PIE-Bench. Ours-short uses approximately one quarter as many sampling steps for prior extraction, while keeping the prior-guided editing stage unchanged. $^\dagger$: results reported in prior work.}
\resizebox{\textwidth}{!}{%
\begin{tabular}{llccccccc}
\toprule
\multirow{2}{*}{\textbf{Method}}
& \multirow{2}{*}{\textbf{Backbone}}
& \multicolumn{1}{c}{\textbf{Structure}}
& \multicolumn{4}{c}{\textbf{Background Preservation}}
& \multicolumn{2}{c}{\textbf{CLIP Similarity}} \\
\cmidrule(lr){3-3} \cmidrule(lr){4-7} \cmidrule(lr){8-9}
& & Distance $\downarrow$
& PSNR $\uparrow$
& LPIPS$_{\times 10^3}$ $\downarrow$
& MSE$_{\times 10^4}$ $\downarrow$
& SSIM$_{\times 100}$ $\uparrow$
& Whole $\uparrow$
& Edited $\uparrow$ \\
\midrule
FlowEdit~\cite{kulikov2025flowedit} & FLUX.1-dev & 28.99 & 21.72 & 104.79 & 98.32 & 83.87 & 25.88 & 22.71 \\
Ours-short & FLUX.1-dev & 7.46 & 30.26 & 21.06 & 19.38 & 94.50 & 25.21 & 22.08 \\
\rowcolor[gray]{0.9} Ours & FLUX.1-dev
& 7.86
& 30.54
& 20.90
& 19.23
& 94.54
& 25.25
& 22.15 \\
\midrule
FlowCycle$^\dagger$~\cite{wang2025flowcycle} & SD3 & 13 & 26.83 & 63 & 30 & 88.6 & 25.48 & 22.46 \\
SplitFlow$^\dagger$~\cite{yoon2025splitflow} & SD3 & 14.55 & 25.22 & 68.53 & 44.96 & 87.54 & 26.23 & 23.01 \\
iRFDS$^\dagger$~\cite{yang2024text} & SD3 & 62.72 & 19.61 & 186.39 & 179.76 & 74.59 & 24.54 & 21.67 \\
FlowAlign~\cite{kim2026flowalign} & SD3 & 36.70 & 24.02 & 65.02 & 56.91 & 86.55 & 26.65 & 22.70\\
UniEdit-Flow~\cite{jiao2026unieditflow} & SD3 & 21.31 & 25.22 & 94.80 & 46.06 & 86.71 & 26.45 & 22.87 \\
FlowEdit~\cite{kulikov2025flowedit} & SD3 & 26.39 & 22.27 & 100.15 & 85.71 & 83.93 & 27.51 & 23.92 \\
Ours-short & SD3 & 10.35 & 27.93 & 32.69 & 28.94 & 91.62 & 26.66 & 23.35 \\
\rowcolor[gray]{0.9} Ours & SD3
& 10.01
& 28.16
& 31.34
& 27.85
& 91.75
& 26.71
& 23.37 \\
\midrule
SplitFlow$^\dagger$~\cite{yoon2025splitflow} & SD3.5 & 11.68 & 27.12 & 52.93 & 30.61 & 89.76 & 26.29 & 22.89 \\
DRFS~\cite{Beaudouin_2026_CVPR} & SD3.5 & 19.48 & 24.36 & 83.09 & 53.85 & 86.15 & 26.55 & 23.72 \\
DNAEdit~\cite{xie2025dnaedit} & SD3.5 & 14.24 & 26.45 & 76.85 & 34.09 & 88.45 & 25.48 & 22.74 \\
FTEdit~\cite{xu2025fteedit} & SD3.5 & 18.40 & 25.23 & 92.36 & 41.59 & 88.77 & 23.96 & 21.17 \\
FlowEdit~\cite{kulikov2025flowedit} & SD3.5 & 22.55 & 23.27 & 88.78 & 69.66 & 85.57 & 27.62 & 23.95 \\
Ours-short & SD3.5 & 9.10 & 28.68 & 29.71 & 25.06 & 91.99 & 26.90 & 23.47 \\
\rowcolor[gray]{0.9} Ours & SD3.5
& 8.90
& 28.78
& 28.79
& 24.59
& 92.03
& 26.86
& 23.35 \\
\bottomrule
\end{tabular}
}
\label{tab:main_results}
\end{table*}

\subsection{CFG Affects Localization and Editing}
\label{subsec:cfg_effects}
\label{subsec:why_decouple}

\begin{figure}[tbp]
\centering
\includegraphics[width=\linewidth]{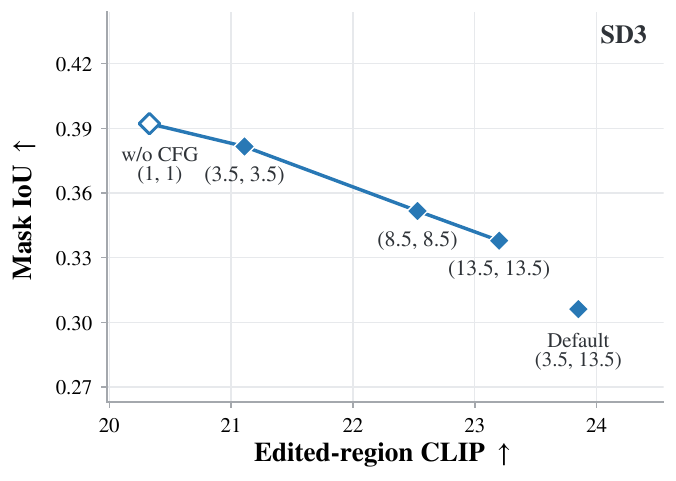}
\caption{CFG trade-off on SD3. Mask IoU is highest without CFG; stronger symmetric CFG raises edited-region CLIP while reducing IoU. Default CFG gives the highest CLIP and lowest IoU.}
\label{fig:cfg_localization}
\end{figure}

We examine how CFG affects localization and editing in FlowEdit to assess DecFlowEdit's two-stage design. We measure localization by IoU between extracted and ground-truth edit masks, and editing fidelity by edited-region CLIP similarity. We test w/o~CFG, FlowEdit's default CFG, and symmetric CFG with guidance scale $g_{\mathrm{src}}=g_{\mathrm{tar}}>1$ (Appendix~\ref{app:cfg_experimental_settings}). Figure~\ref{fig:cfg_localization} shows a guidance-dependent trade-off consistent with Equation~\eqref{eq:cfg_decomposition}. Without CFG removes both additional terms, yielding the highest IoU but the lowest CLIP. Default CFG retains both terms and achieves the highest CLIP but the lowest IoU. Symmetric CFG removes the asymmetry term, but increasing the common guidance scale raises CLIP while reducing IoU.

\begin{figure*}[tbp]
\centering
\includegraphics[width=\textwidth]{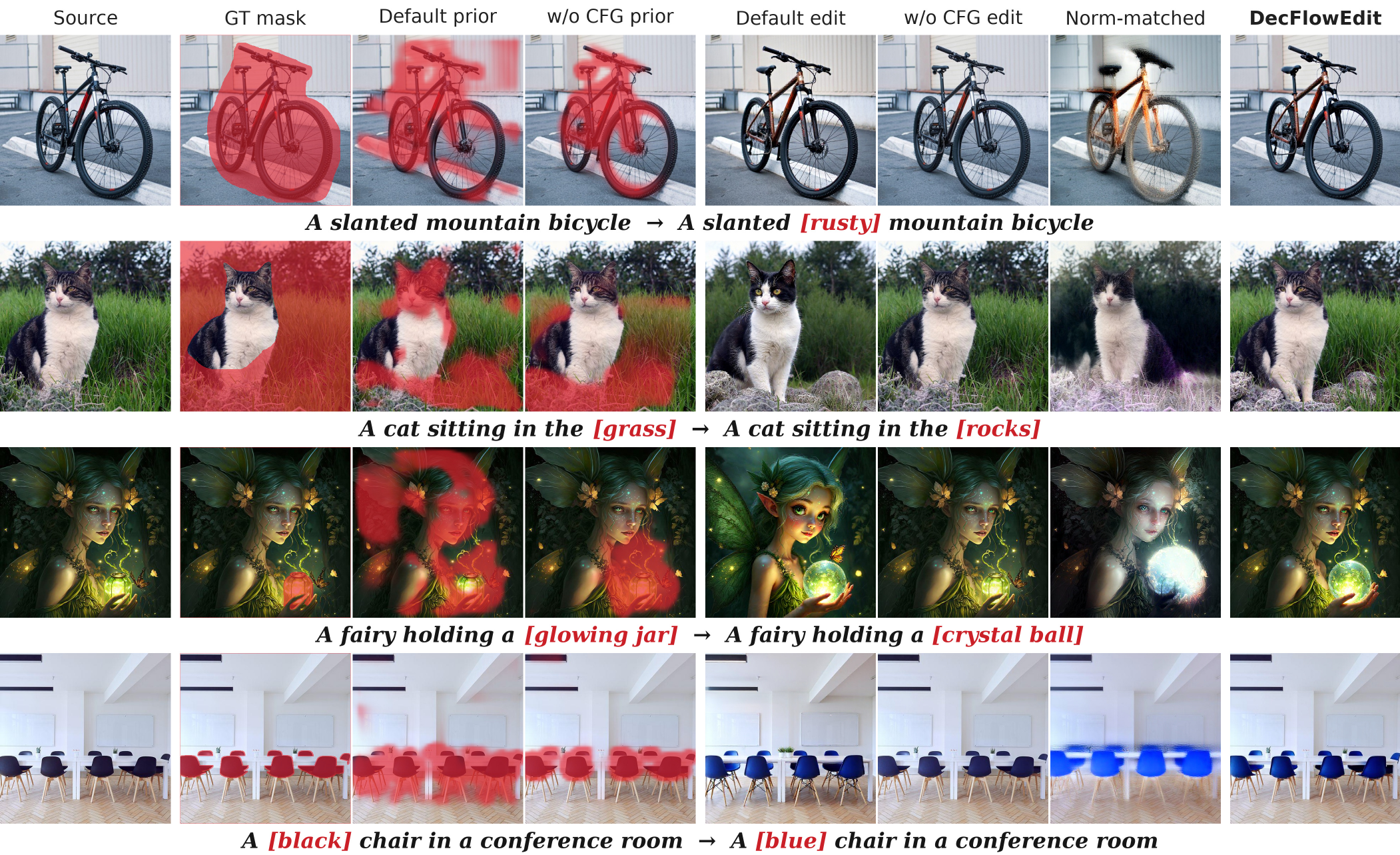}
\caption{Qualitative comparison of ground-truth masks, extracted priors, and editing results from FlowEdit under w/o~CFG and default CFG, together with norm-matched editing and DecFlowEdit, all using FLUX.1-dev. }
\label{fig:decouple_qualitative}
\end{figure*}

We further measure the regional norms of the decomposition terms along the actual trajectories of five SD3 CFG configurations on 556 PIE-Bench local-edit tasks (Appendix~\ref{app:cfg_decomposition_empirical}). Under default CFG, the asymmetry term has a large magnitude and weaker edit/non-edit contrast than the first term. Under symmetric CFG, the amplification term also has weaker contrast and grows relative to the first term as guidance increases, while the total non-edit-to-edit response ratio rises. Guidance also changes the trajectory states, and the first term itself exhibits weaker contrast along strongly guided trajectories. The total response ratio is lowest without CFG. Together with Figures~\ref{fig:guidance_observation} and~\ref{fig:decouple_qualitative}, these results support extracting localization information along a separate w/o~CFG trajectory to constrain editing under default CFG.
We further compare priors extracted under w/o~CFG and default CFG, keeping the editing stage fixed at default CFG. Priors extracted under w/o~CFG reduce background LPIPS by approximately 39--50\% relative to those extracted under default CFG, while edited-region CLIP differs by less than 0.13\%. These results show that the localization advantage of w/o~CFG translates into better background preservation in the final edit. Experimental settings and detailed results are provided in Appendix~\ref{app:cfg_backbones}.

\par

\paragraph{Global amplification of w/o~CFG updates.}

We test whether amplifying w/o~CFG updates can recover default CFG's editing performance. At each step, we compute both updates at the same paired source/target states with the same sampled noise, then globally scale the w/o~CFG update to match the default CFG update's Frobenius norm. The norm-matched outputs in column 7 of Figure~\ref{fig:decouple_qualitative} exhibit severe blur and loss of detail. This approach yields lower edited-region CLIP and higher structure distance than default CFG across FLUX.1-dev, SD3, and SD3.5 (Appendix~\ref{app:global_norm_matching}).

\subsection{Guidance-dependent Localization}
\label{subsec:guidance_localization_analysis}

We analyze the localization advantage of w/o~CFG through spatial and temporal differences between edit and non-edit regions. FlowEdit's shared-noise construction favors cancellation of common content between the source and target velocities (Appendix~\ref{app:within_step_analysis}). In the goat-to-horse example in Figure~\ref{fig:guidance_observation}, both prompts describe the same neighboring cat, but default CFG uses different guidance strengths and can produce different velocities even for this unchanged object. Their difference can therefore contain unintended changes. Without CFG removes this guidance mismatch, favoring cancellation in non-edit regions.

\paragraph{Within-step spatial contrast.}
We compare unmasked FlowEdit trajectories on PIE-Bench, using ground-truth masks mapped to the latent grid to define the edit region $\Omega_E$ and its complement, the non-edit region $\Omega_N$. At each update step, $R_t$ is the ratio of the non-edit-region to edit-region mean velocity-difference magnitude. Averaging over update steps within each sample and then across samples gives $0.5175$ under w/o~CFG and $0.8007$ under default CFG on FLUX.1-dev. Thus, non-edit-region magnitudes are roughly half the edit-region magnitudes under w/o~CFG, whereas default CFG makes the regions harder to distinguish. SD3 and SD3.5 show the same trend (Appendix~\ref{app:cfg_spatial_contrast}).

\paragraph{Temporal directional contrast.}
Beyond the within-step spatial contrast, we examine how velocity differences in edit and non-edit regions combine during temporal aggregation in Equation~\eqref{eq:without_cfg_energy}. More consistent directions favor accumulation in the vector sum. We therefore flatten each region's velocity difference into $\mathbf{u}^{\Omega}_t=\operatorname{vec}(\Delta\mathbf{V}_t|_{\Omega})$ and measure the mean cosine similarity over all distinct timestep pairs in $\mathcal{T}_{\mathrm{mid}}=\{t_1,\ldots,t_m\}$, where $m=|\mathcal{T}_{\mathrm{mid}}|$ is the number \mbox{of selected timesteps:}
\begin{equation}
\mathrm{Cos}_{\Omega}
=\frac{2}{m(m-1)}\sum_{i<j}
\frac{\langle\mathbf{u}^{\Omega}_{t_i},\mathbf{u}^{\Omega}_{t_j}\rangle}
{\|\mathbf{u}^{\Omega}_{t_i}\|_2\,\|\mathbf{u}^{\Omega}_{t_j}\|_2}.
\label{eq:cfg_temporal_cosine}
\end{equation}
Averaged over PIE-Bench editing tasks performed with FLUX.1-dev, the edit-region Cos exceeds the non-edit-region Cos by a larger margin under w/o~CFG than under default CFG ($0.12740$ vs.\ $0.04984$). Together with the cancellation analysis in Appendix~\ref{app:cfg_temporal_contrast}, this indicates that temporal aggregation under w/o~CFG more selectively retains edit-region velocity differences while attenuating residuals in non-edit regions.

\par
\begin{table*}[!tbp]
\centering
\caption{Localization quality and downstream editing performance on the same 490 PIE-Bench local-edit tasks (Appendix~\ref{app:prior_comparison_protocol}). Metrics: IoU~$\uparrow$, edited-region CLIP~$\uparrow$, background LPIPS ($\times 10^3$)~$\downarrow$, and structure distance~$\downarrow$.}
\label{tab:spatial_prior_comparison}
\small
\setlength{\tabcolsep}{4pt}
\begin{tabular*}{\textwidth}{@{\extracolsep{\fill}}llrrrr@{}}
\toprule
Source & Model for prior extraction & IoU & CLIP & LPIPS & Structure \\
\midrule
DiffEdit-Single & FLUX.1-dev & 0.3392 & 20.860 & 30.119 & 8.351 \\
DiffEdit-Multi & FLUX.1-dev & 0.3678 & 21.075 & 36.842 & 10.849 \\
MasaCtrl & FLUX.1-dev & 0.3288 & 21.392 & 58.935 & 16.430 \\
MasaCtrl & SD1.4 & 0.3087 & 21.071 & 55.702 & 14.903 \\
Prompt-to-Prompt & SD1.4 & 0.3436 & 21.588 & 69.940 & 20.707 \\
Auto VLM$\rightarrow$SAM3 & Qwen3-VL-8B + SAM3 & 0.5493 & 20.857 & 21.286 & 8.620 \\
\textbf{Ours} & FLUX.1-dev & 0.4195 & 21.290 & 21.190 & 7.978 \\
\bottomrule
\end{tabular*}
\end{table*}

\subsection{Comparison of Spatial Priors}
\label{subsec:spatial_prior_comparison}

To evaluate the spatial constraints from prior extraction, we compare our prior with localization signals from diffusion-based editing pipelines~\cite{couairon2023diffedit,cao2023masactrl,hertz2023p2p} and an external segmentation pipeline. The external pipeline uses Qwen3-VL-8B~\cite{bai2025qwen3vl} as a vision-language model (VLM) to generate a query from the source image and editing instruction, followed by SAM3~\cite{carion2025sam3} segmentation. Table~\ref{tab:spatial_prior_comparison} reports IoU against ground-truth edit masks and downstream editing and preservation metrics. We provide evaluation details and mask extraction procedures for all baselines in Appendix~\ref{app:mask_extraction}.

\paragraph{Internal localization signals.} DecFlowEdit obtains spatial information by aggregating velocity differences under w/o~CFG during prior extraction. Compared with the other internal localization methods in Table~\ref{tab:spatial_prior_comparison}, our prior achieves higher IoU, lower background LPIPS, and lower structure distance, while retaining competitive edited-region CLIP. These results show that temporally aggregated velocity differences provide useful localization cues and effective spatial constraints for prior-guided editing.

\paragraph{External segmentation.} In the end-to-end comparison with Auto VLM$\rightarrow$SAM3, DecFlowEdit achieves higher edited-region CLIP and lower structure distance at similar background LPIPS, despite lower mask IoU. Our prior is intended to capture regions affected by editing. In the rusty-bicycle example in Figure~\ref{fig:decouple_qualitative}, it highlights metal components such as the frame and rims rather than the entire object.

\section{Conclusion}
\label{sec:conclusion}
We present DecFlowEdit, a training-free and inversion-free method that improves spatial control in flow-based image editing. Our analysis reveals distinct roles of guidance in localization and semantic editing: w/o~CFG favors localization, whereas default CFG favors effective semantic editing. Exploiting this distinction, DecFlowEdit better preserves edit-irrelevant content without modifying the generative model or requiring additional supervision. Experiments across three flow-based backbones show that this simple separation yields more reliable editing trajectories and a better balance between semantic fidelity and structure preservation. More broadly, our results suggest that internal dynamics of the flow-based image editing can provide useful spatial signals for edit control.
\subsection*{LLM Usage Statement}
We used OpenAI Codex to assist with writing and language polishing, identifying and retrieving related literature, and research ideation and execution, including code implementation, debugging, and preparation of figures and tables. The authors take responsibility for the final manuscript and accompanying artifacts.

{
\small
\bibliographystyle{ieeenat_fullname}
\bibliography{main}
}


\clearpage
\appendix
\section{Additional Analysis of Guidance-dependent Localization}
\label{app:guidance_localization_analysis}

This appendix supplements Section~\ref{subsec:guidance_localization_analysis} with a conceptual explanation, spatial-contrast measurements, and temporal directional and cancellation analysis.

\subsection{Intuition for Guidance-dependent Localization}
\label{app:within_step_analysis}

FlowEdit evaluates source and target velocities at paired states that share the same sampled noise. When the editing state stays close to the source image in a non-edit region, the paired states are locally similar, allowing velocity components associated with common content to partially cancel in their difference. Shared semantics alone, however, do not ensure identical velocities: default CFG applies different source and target guidance strengths, which can leave residual updates even on unchanged objects. In Figure~\ref{fig:guidance_observation}, both prompts preserve the cat beside the goat, yet the guidance mismatch can produce different velocities on the cat. Without CFG substantially reduces this mismatch, favoring cancellation in non-edit regions. The spatial and temporal measurements in Section~\ref{subsec:guidance_localization_analysis} quantify differences between edit and non-edit regions under the two CFG settings.

\begin{figure*}[!htbp]
    \centering
    \includegraphics[width=0.85\textwidth]{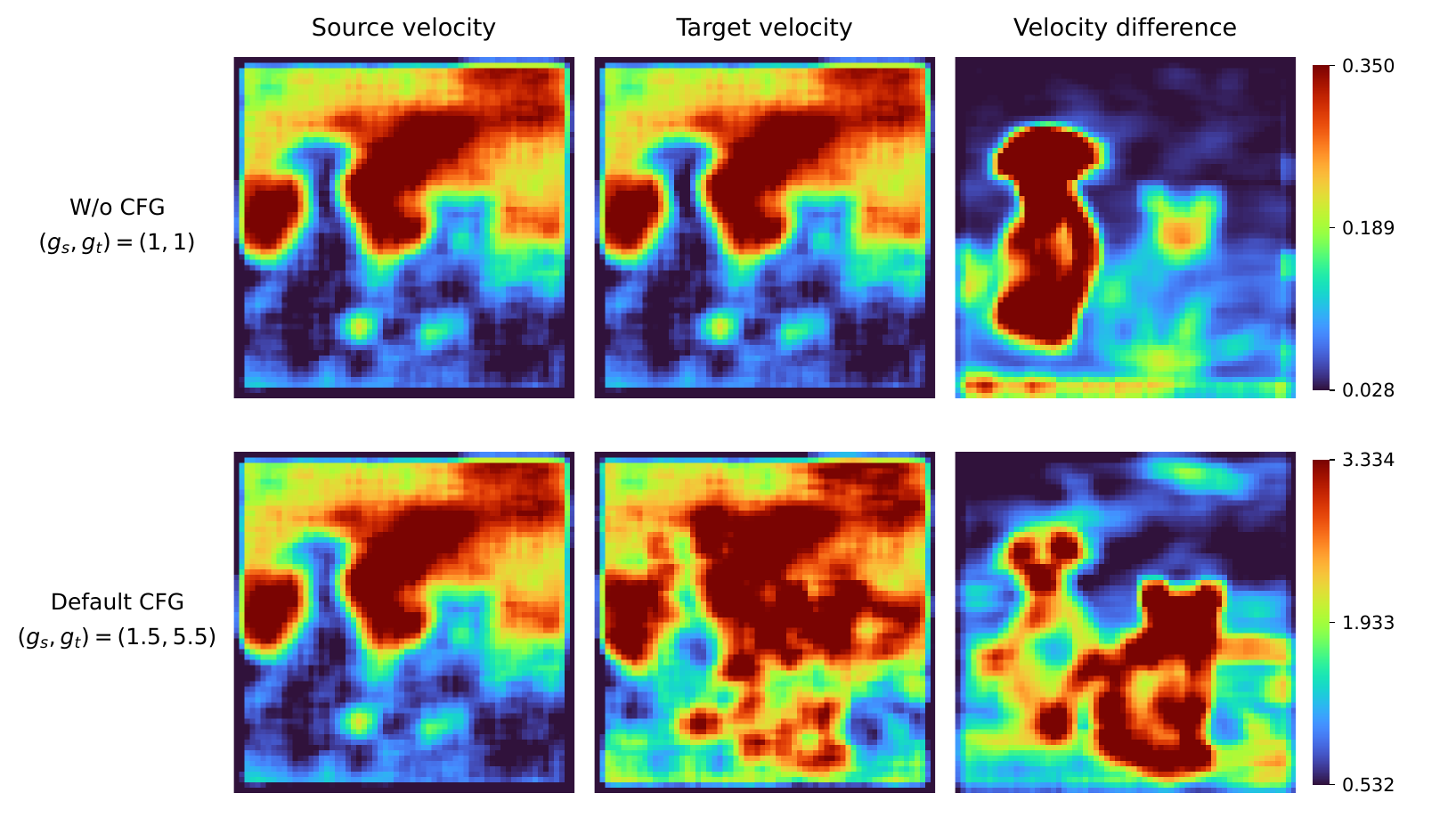}
    \caption{Magnitudes of the source velocity, target velocity, and their vector difference, evaluated at the same noisy source state at Step 15 under w/o~CFG and default CFG (FLUX.1-dev).}
    \label{fig:cfg_velocity_step15}
\end{figure*}

For the goat-to-horse example in Figure~\ref{fig:guidance_observation}, Figure~\ref{fig:cfg_velocity_step15} evaluates both prompt branches at the same noisy source state at Step 15 of the 28-step FLUX.1-dev schedule. We take the $\ell_2$ norm over channels and smooth each magnitude map with a $5\times5$ mean filter. The four source/target maps share the pooled 10th--90th-percentile range $[3.616,7.156]$, while each velocity-difference map uses its own percentile range.

\subsection{Within-step Spatial Contrast}
\label{app:cfg_spatial_contrast}

We detail the spatial-contrast experiment in Section~\ref{subsec:guidance_localization_analysis} on PIE-Bench tasks with nonempty edit and non-edit regions. Ground-truth edit masks are resized to the $64\times64$ latent grid by nearest-neighbor interpolation to define the edit region $\Omega_E$; its complement is the non-edit region $\Omega_N$. We compare unmasked FlowEdit trajectories under w/o~CFG $(1,1)$ and default CFG: $(1.5,5.5)$ for FLUX.1-dev and $(3.5,13.5)$ for SD3 and SD3.5. Within each backbone, both configurations use the same inputs, sampling schedule, and paired noise draws, while their states evolve separately.

Let $r_t(x)=\|\Delta\mathbf{V}_t(x)\|_2$ denote the velocity-difference magnitude at location $x$, with the norm taken over channels. We compute the ratio of non-edit-region to edit-region mean magnitudes:
\begin{equation}
R_t
=\frac{|\Omega_N|^{-1}\sum_{x\in\Omega_N}r_t(x)}
{|\Omega_E|^{-1}\sum_{x\in\Omega_E}r_t(x)}.
\label{eq:keep_edit_response_ratio}
\end{equation}
For each sample, we average $R_t$ over all actual update steps: indices 4--27 of FLUX.1-dev's 28-step schedule and indices 17--49 of the 50-step SD3 and SD3.5 schedules, using zero-based indexing. We then average these per-sample values equally to obtain $\overline R$. The calculation uses raw velocity differences, without percentile normalization or temporal aggregation.

Switching from w/o~CFG to default CFG increases $\overline R$ from $0.5175$ to $0.8007$ on FLUX.1-dev, from $0.5985$ to $0.8334$ on SD3, and from $0.5903$ to $0.8409$ on SD3.5. The sample-averaged $R_t$ is higher under default CFG at every actual update step: 24/24 on FLUX.1-dev and 33/33 on each SD backbone. Thus, w/o~CFG produces a clearer magnitude contrast between edit and non-edit regions across all three backbones. These compare the evolving trajectories; the per-step trend refers to sample means.

\subsection{Temporal Directional Contrast}
\label{app:cfg_temporal_contrast}

Section~\ref{subsec:guidance_localization_analysis} finds a larger gap in temporal directional consistency between edit and non-edit regions under w/o~CFG. Here we detail the cosine computation and examine whether this directional contrast is accompanied by stronger relative cancellation of velocity differences in non-edit regions during temporal aggregation.

We use the FLUX.1-dev trajectories and region definitions described above. Following Equation~\eqref{eq:cfg_temporal_cosine}, $\mathbf{u}^{\Omega}_t=\operatorname{vec}(\Delta\mathbf{V}_t|_{\Omega})$ concatenates all spatial positions and channels within region $\Omega$. The intermediate window $\mathcal{T}_{\mathrm{mid}}$ contains indices 7--21 of the full 28-step schedule. Cos averages the cosine similarities over all distinct timestep pairs in this window. To measure cancellation when these velocity differences are aggregated as in Equation~\eqref{eq:without_cfg_energy}, we additionally compute the temporal cancellation ratio:
\begin{equation}
\mathrm{TCR}_{\Omega}
=\frac{\left\|\sum_{t\in\mathcal{T}_{\mathrm{mid}}}\mathbf{u}^{\Omega}_t\right\|_2}
{\sum_{t\in\mathcal{T}_{\mathrm{mid}}}\|\mathbf{u}^{\Omega}_t\|_2+\delta},
\label{eq:cfg_temporal_cancellation_ratio}
\end{equation}
where $\delta=10^{-8}$. The raw vectors are summed with equal weights, without multiplying by timestep sizes. Lower TCR indicates that a smaller fraction of the summed per-step magnitudes remains after vector addition, corresponding to stronger cancellation. Unlike Cos, TCR depends on both directions and magnitudes. Both metrics are computed per sample and then averaged equally across samples; $\Delta$ denotes the edit-region value minus the non-edit-region value.

\begin{table*}[!htbp]
\centering
\caption{Temporal directional contrast and cancellation on FLUX.1-dev. Cos and TCR use the same intermediate steps, indices 7--21. E and N denote the edit and non-edit regions; $\Delta$ denotes E minus N.}
\label{tab:cfg_temporal_contrast}
\small
\setlength{\tabcolsep}{5pt}
\begin{tabular}{lcccccc}
\toprule
\multirow{2}{*}{CFG} & \multicolumn{3}{c}{Cos} & \multicolumn{3}{c}{TCR} \\
\cmidrule(lr){2-4}\cmidrule(lr){5-7}
& E & N & $\Delta\,\uparrow$ & E & N & $\Delta\,\uparrow$ \\
\midrule
w/o~CFG & 0.53305 & 0.40565 & \textbf{0.12740} & 0.74017 & 0.65547 & \textbf{0.08470} \\
default CFG & 0.55259 & 0.50275 & 0.04984 & 0.75860 & 0.72725 & 0.03135 \\
\bottomrule
\end{tabular}
\end{table*}

Table~\ref{tab:cfg_temporal_contrast} shows larger gaps between edit and non-edit regions under w/o~CFG than under default CFG in both Cos ($0.12740$ vs.\ $0.04984$) and TCR ($0.08470$ vs.\ $0.03135$). These larger gaps indicate stronger contrast between the two regions in directional consistency and in the fraction of cumulative velocity-difference magnitude retained after vector aggregation. This stronger contrast helps temporal aggregation distinguish edit regions from non-edit regions, supporting the use of w/o~CFG for prior extraction.


\section{Mask Extraction Baselines}
\label{app:mask_extraction}

\begin{figure*}[t]
    \centering
    \includegraphics[width=1.00\linewidth]{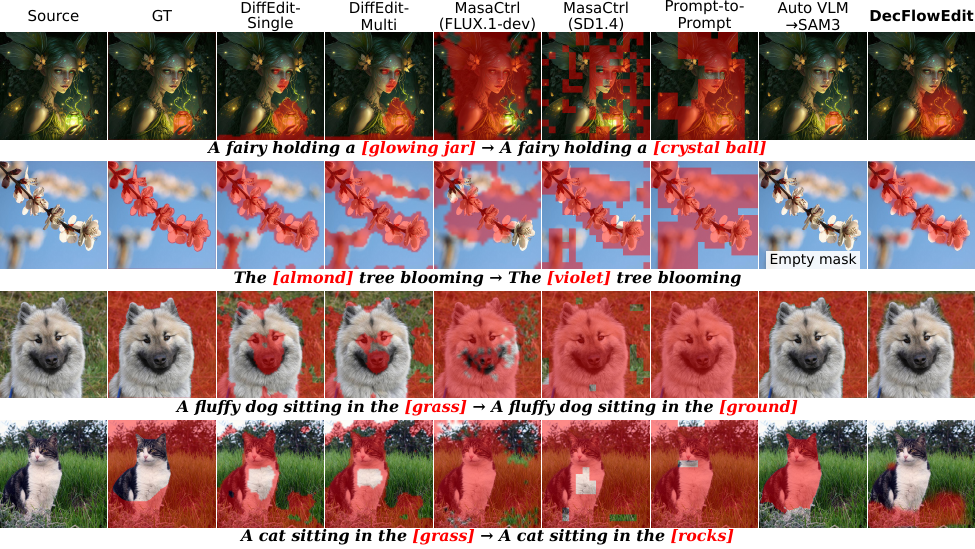}
    \caption{Additional qualitative comparison of different localization masks. In the second row, SAM3 returns no instance above the confidence threshold ($0.5$) for the VLM-generated query ``almond tree'', yielding an empty mask. GT: ground-truth edit mask. }
    \label{fig:mask2}
\end{figure*}

We describe the FLUX.1-dev-based baselines, SD1.4 attention-control masks, and the external Auto VLM$\rightarrow$SAM3 pipeline compared in Table~\ref{tab:spatial_prior_comparison}.

\paragraph{DiffEdit-Single and DiffEdit-Multi.}
Both variants localize edits using source--target velocity disagreement on noisy versions of the source image. Given the source latent $\mathbf{x}^{\mathrm{src}}$ and noise samples $\boldsymbol{\epsilon}_i\sim\mathcal{N}(\mathbf{0},\mathbf{I})$, we construct $\mathbf{z}_{t_j}^{(i)}=(1-\sigma_j)\mathbf{x}^{\mathrm{src}}+\sigma_j\boldsymbol{\epsilon}_i$ and compute
\[
d^{(i,j)}(h,w)
=\left\|\begin{gathered}
\mathbf{V}_\theta(\mathbf{z}_{t_j}^{(i)},t_j\mid P_{\mathrm{tar}})(:,h,w)\\
-\mathbf{V}_\theta(\mathbf{z}_{t_j}^{(i)},t_j\mid P_{\mathrm{src}})(:,h,w)
\end{gathered}\right\|_2.
\]
The localization map averages these magnitude maps over $k$ timesteps and $n$ noise samples:
\[
\bar d(h,w)=\frac{1}{kn}\sum_{j=1}^{k}\sum_{i=1}^{n}d^{(i,j)}(h,w).
\]
DiffEdit-Single uses $k=1$, selecting a single noise level with strength $0.5$ from a $T=28$ schedule. DiffEdit-Multi uses $k=5$ timesteps uniformly selected from $\sigma\in[0.2,0.8]$. Both use $n=10$ and source/target guidance scales of $(1,1)$. We apply min--max normalization to $\bar d$ before binarization.

Our baselines adapt DiffEdit to FLUX.1-dev, whereas the original work uses class-conditional latent diffusion and text-conditional Stable Diffusion models~\cite{couairon2023diffedit}. On 556 local-edit tasks, applying the original default threshold of $0.5$ to our FLUX.1-dev implementation yields low IoUs of $0.0702$ and $0.1228$ for DiffEdit-Single and DiffEdit-Multi, respectively. We therefore sweep the threshold and adopt $\tau=0.15$ for both variants (IoUs of $0.3368$ and $0.3615$ on the same subset).

\paragraph{MasaCtrl (SD1.4 and FLUX.1-dev).}
\emph{SD1.4.} We use $512\times512$ images, source-image DDIM inversion, and 50-step DDIM sampling with guidance scale $7.5$. PIE-Bench blend words identify the source and target control tokens. MasaCtrl~\cite{cao2023masactrl} collects $16\times16$ cross-attention maps within each step, aggregates the selected-token maps over heads and layers, and applies min--max normalization to each branch. The attention store is reset after each step. Mutual self-attention control starts at step 4 and block 10 (zero-based). The target map is resized by nearest-neighbor interpolation to the current attention resolution and binarized using the official default threshold of $0.1$. We export the target mask from the last control application at the final sampling step.

\emph{FLUX.1-dev.} The MasaCtrl-style mask is extracted from joint attention maps in the FLUX.1-dev double-stream transformer blocks. We first locate the token indices corresponding to the edited word in the source prompt. During the forward pass, we extract the cross-attention submatrix from the source text tokens to image patches:
\[
\mathrm{Attn}_{\mathrm{edit}}
=
\mathrm{Attn}
[:, :, \mathcal{I}_{\mathrm{token}}, S_{\mathrm{text}}:],
\]
where \(\mathcal{I}_{\mathrm{token}}\) denotes the token indices of the edited word and \(S_{\mathrm{text}}\) is the text sequence length. We average the attention maps over tokens, heads, layers, and timesteps to obtain a spatial attention map. The map is min--max normalized to \([0,1]\) and binarized at \(\tau=0.1\), using the same threshold as the official SD1.4 implementation. We use \(T=28\) and guidance scale \(3.5\).

These FLUX.1-dev-based baselines represent different localization signals: DiffEdit-Single uses single-step source--target velocity disagreement, DiffEdit-Multi uses multi-step velocity-disagreement aggregation, and the MasaCtrl-style mask uses internal attention signals. In contrast, our spatial prior is derived directly from temporally aggregated FlowEdit velocity differences, making it better aligned with the editing operator used for subsequent reweighting.

\paragraph{Prompt-to-Prompt (SD1.4).}
We use $512\times512$ images, source-image DDIM inversion, and 50-step DDIM sampling with guidance scale $7.5$. PIE-Bench blend words identify the source and target control tokens. We use the official attention-refinement controller~\cite{hertz2023p2p}, with cross-attention and self-attention replacement fractions of $0.8$ and $0.4$. LocalBlend uses the $16\times16$ maps from \texttt{down\_cross[2:4]} and \texttt{up\_cross[:3]}, accumulated over all 50 sampling steps. It aggregates the selected-token maps over heads and layers, applies $3\times3$ max-pooling and nearest-neighbor interpolation, and divides each branch's map by its spatial maximum. The source and target maps are binarized using LocalBlend's official default threshold of $0.3$ and combined by union. We export the hard mask used by LocalBlend at the final step.

\paragraph{Auto VLM$\rightarrow$SAM3.}
Qwen3-VL-8B-Instruct~\cite{bai2025qwen3vl} receives the source image, source and target descriptions, and editing instruction, and returns a single source-side object or part phrase. We use greedy decoding with at most 32 generated tokens. SAM3~\cite{carion2025sam3} segments the source image using this phrase, with its image processor resizing the input to $1008\times1008$. We retain all instances with confidence above the official default threshold of $0.5$, resize their mask logits to the source-image resolution, and apply a sigmoid. For IoU evaluation, we binarize each instance using the official pixel-probability threshold of $0.5$ and take the union. For downstream editing, we instead take the pixel-wise maximum of the retained instance probabilities.

\paragraph{Evaluation protocol for the spatial-prior comparison.}
\label{app:prior_comparison_protocol}
All metrics in Table~\ref{tab:spatial_prior_comparison} are evaluated on the same 490 local-edit tasks in PIE-Bench. Starting from its 700 tasks, we exclude 144 whose ground-truth masks cover the entire image and 66 for which the SD1.4 controllers cannot identify the required source--target token pair (46 with unparseable blend-word pairs and 20 with no exact token match). Mask extraction succeeds for both SD1.4 methods on the remaining $700-144-66=490$ tasks. The Model column identifies the model used to produce each prior; all priors guide FLUX.1-dev editing under default CFG. IoU is measured against the ground-truth edit masks. Edited-region CLIP is scaled by 100, and background LPIPS and structure distance by 1000. For Auto VLM$\rightarrow$SAM3, if the extracted mask is empty, as in the second row of Figure~\ref{fig:mask2}, we use an all-one weight map. The SD1.4 binary masks and SAM3 soft probability maps are saved at source-image resolution; SAM3 probabilities are stored as 8-bit grayscale values. For editing, the maps are converted to $[0,1]$ and resized bilinearly to the FLUX.1-dev latent resolution, without further thresholding, then used directly as the spatial update weights.

\section{Implementation Details}
\label{app:implementation_details}

\paragraph{Backbones and environment.}
We evaluate DecFlowEdit on FLUX.1-dev, Stable Diffusion 3 (SD3), and Stable Diffusion 3.5-Medium (SD3.5), following the standard FlowEdit settings for each backbone. Runtime measurements use a single NVIDIA A100-SXM4-80GB GPU under Ubuntu~24.04.3, with the CUDA~12.1 runtime. FLUX.1-dev and SD3 use Python~3.8.20, PyTorch~2.4.1+cu121, and diffusers~0.30.1; SD3.5 uses Python~3.10.20, PyTorch~2.5.1+cu121, and diffusers~0.37.1.

\begin{table*}[t]
\centering
\caption{NFE counts and measured end-to-end runtimes on PIE-Bench. Times are mean $\pm$ sample standard deviation over 10 images at $512\times512$ resolution on one A100-SXM4-80GB GPU. Parenthesized factors are NFE ratios relative to FlowEdit on the same backbone. Peak memory is the maximum allocated GPU memory over the measured images, including model weights.}
\label{tab:compute_resources}
\small
\setlength{\tabcolsep}{4pt}
\begin{tabular}{@{}llccc@{}}
\toprule
\textbf{Backbone} & \textbf{Method} & \textbf{NFEs (ratio)} & \textbf{Time (s)} & \shortstack{\textbf{Peak memory}\\\textbf{(GiB)}} \\
\midrule
FLUX.1-dev & FlowEdit & 48 ($1.00\times$) & $7.882\pm0.016$ & 34.17 \\
& Ours & 96 ($2.00\times$) & $15.762\pm0.005$ & 34.17 \\
& Ours-short & 60 ($1.25\times$) & $10.066\pm0.017$ & 34.17 \\
\midrule
SD3 & FlowEdit & 132 ($1.00\times$) & $2.705\pm0.137$ & 17.22 \\
& Ours & 198 ($1.50\times$) & $4.094\pm0.056$ & 17.00 \\
& Ours-short & 150 ($1.14\times$) & $3.199\pm0.072$ & 17.00 \\
\midrule
SD3.5 & FlowEdit & 132 ($1.00\times$) & $3.761\pm0.012$ & 17.92 \\
& Ours & 198 ($1.50\times$) & $5.789\pm0.017$ & 17.92 \\
& Ours-short & 150 ($1.14\times$) & $4.423\pm0.019$ & 17.92 \\
\bottomrule
\end{tabular}
\end{table*}

\paragraph{Sampling schedules and CFG.}
The prior-guided editing stage uses 28 sampling steps for FLUX.1-dev and 50 for SD3/SD3.5. Its default CFG configuration $\mathbf{g}=(g_{\mathrm{src}},g_{\mathrm{tar}})$ is $(1.5,5.5)$ for FLUX.1-dev and $(3.5,13.5)$ for SD3/SD3.5. The prior-extraction stage uses $\mathbf{g}^{\mathrm{w/oCFG}}=(1,1)$. The full variant uses the same number of sampling steps for both stages, whereas Ours-short uses approximately one quarter as many steps for prior extraction while keeping the prior-guided editing stage unchanged.

\paragraph{NFE counts and runtime.}
We count each conditional or unconditional velocity prediction as one function evaluation (NFE), including predictions evaluated together in a batched transformer call. FLUX.1-dev uses guidance distillation: each step evaluates the source and target velocities through two separate transformer calls, giving 2 NFEs in both prior extraction and editing. SD3/SD3.5 use explicit CFG during editing, evaluating the source and target conditional and unconditional predictions in one batched call, giving 4 NFEs per step. During prior extraction, $g_{\mathrm{src}}=g_{\mathrm{tar}}=1$ yields $V^g=V^u+g(V^c-V^u)=V^c$, where $V^c$ and $V^u$ denote conditional and unconditional velocities. For these measurements, we therefore evaluate only the two conditional predictions in one batched call, giving 2 NFEs per prior-extraction step.

FLUX.1-dev executes 24 editing steps and 24/6 prior-extraction steps for the full/short variants; SD3/SD3.5 execute 33 editing steps and 33/9 prior-extraction steps. The resulting NFE counts for FlowEdit, Ours, and Ours-short are $48$, $48+24\times2=96$, and $48+6\times2=60$ on FLUX.1-dev, and $132$, $132+33\times2=198$, and $132+9\times2=150$ on SD3/SD3.5. Table~\ref{tab:compute_resources} reports these counts and measured end-to-end runtimes. Each setting uses two warm-up images followed by the same 10 PIE-Bench images for timing. Timings include image loading and preprocessing, VAE and prompt encoding, prior extraction and mask construction when applicable, editing, and decoding, with CUDA synchronization at stage boundaries; model loading and output-file writing are excluded.

\paragraph{Spatial-prior construction and settings.}
We construct the spatial prior from the energy map $E(x)$ defined in
Section~\ref{subsec:without_cfg_prior}. We first apply a $5\times5$ mean filter
$S$, obtaining $A(x)=S(E)(x)$. To reduce sensitivity to the response scale
across samples, we normalize the smoothed map using its spatial percentiles:
\begin{equation}
\widetilde{E}(x)
=
\operatorname{clip}
\left(
\frac{A(x)-q_{10}}
{\max\{q_{90}-q_{10},\delta\}},
0,1
\right),
\qquad
\delta=10^{-8},
\label{eq:prior_normalization}
\end{equation}
where $q_{10}$ and $q_{90}$ are the 10th and 90th percentiles of all spatial
values in $A$ for the current sample, and $\operatorname{clip}(\cdot,0,1)$
clips its input to $[0,1]$.

The normalized response is converted into a soft spatial prior:
\begin{equation}
M(x)
=
\sigma
\left(
\frac{\widetilde{E}(x)-\tau}{s}
\right),
\qquad
s>0,
\label{eq:soft_spatial_prior}
\end{equation}
where $\sigma$ is the sigmoid function, and $\tau$ and $s$ denote the threshold
and softness parameters. We use the same defaults, $\tau=0.55$ and $s=0.08$, across FLUX.1-dev, SD3, and SD3.5. For the full prior-extraction schedule, all three backbones follow the intermediate-window rule defined in Section~\ref{subsec:without_cfg_prior}. We adopt $\tau=0.55$ as a preservation-oriented operating point.

We evaluate parameter sensitivity on PIE-Bench, using all three backbones for the threshold analysis and FLUX.1-dev for the softness and aggregation-window analyses. The threshold and window experiments evaluate edited-region CLIP on all 700 tasks, and localization and background-preservation metrics on the 556 local-edit tasks. The downstream softness sweep evaluates all metrics on these 556 tasks. Appendix~\ref{app:tau_sensitivity}--\ref{app:aggregation_window_sensitivity} provides the detailed results, showing the editing--preservation trade-offs and the robustness to the aggregation-window choice.

For Ours-short, the prior-extraction stage uses 7 scheduled steps for FLUX.1-dev and 13 for SD3/SD3.5. We aggregate velocity differences at zero-based indices 2--5 for FLUX.1-dev and 4--9 for SD3/SD3.5, including both endpoints. For mask-quality evaluation, we binarize the soft spatial prior $M$ at $0.5$. The prior-guided editing stage uses the soft $M$ throughout, following Section~\ref{subsec:default_cfg_editing}.

\section{Parameter Sensitivity and Editing--Preservation Trade-offs}
\label{app:parameter_sensitivity}
\label{app:edit_preservation_tradeoff}

We examine how update strength, the spatial-prior threshold $\tau$, softness $s$, and the aggregation window affect localization and final editing. The first three controls change the amount or spatial distribution of the applied update; the window determines which responses form the prior. Unless specified otherwise, we use $\tau=0.55$ and $s=0.08$, and retain each backbone's default CFG in the prior-guided editing stage. These experiments characterize trade-offs rather than identify one setting that maximizes every metric.

\subsection{Spatial Reweighting versus Uniform Scaling}
\label{app:spatial_uniform_control}

We test whether the background-preservation gains in Table~\ref{tab:main_results} can be explained solely by weaker editing. To assess the benefit of spatial selectivity, we compare spatial reweighting with uniform update scaling at comparable edited-region CLIP.

\paragraph{Control formulas.}
Let $\Delta\mathbf{V}^{\mathrm{def}}_t(x)$ denote the velocity difference under default CFG and $M(x)\in[0,1]$ the fixed soft spatial prior for an image. We compare
\begin{align}
\Delta\mathbf{V}^{\mathrm{Dec},\lambda}_t(x)
&=\big[\lambda+(1-\lambda)M(x)\big]\Delta\mathbf{V}^{\mathrm{def}}_t(x),
\label{eq:spatial_control_sweep}\\
\Delta\mathbf{V}^{\mathrm{uniform},\lambda}_t(x)
&=\lambda\,\Delta\mathbf{V}^{\mathrm{def}}_t(x).
\label{eq:uniform_control_sweep}
\end{align}
Both controls recover raw FlowEdit at $\lambda=1$. At $\lambda=0$, spatial reweighting recovers the DecFlowEdit update in Equation~\eqref{eq:decflowedit_update}, whereas uniform scaling supplies no editing update. Decreasing $\lambda$ therefore strengthens suppression in both controls, but spatial reweighting targets low-prior locations while uniform scaling attenuates every location and channel equally. We compare the controls at comparable CLIP, not at equal $\lambda$. Each setting evaluates the velocity difference along its own trajectory.

\paragraph{Experimental settings and evaluation.}
Both controls use $\lambda\in\{0,0.2,0.4,0.6,0.8,1\}$ on PIE-Bench. We fix FLUX.1-dev, default CFG $(1.5,5.5)$, a 28-step schedule with 24 actual updates, and one fresh paired noise sample per step. The prior built from the aggregated response $E$ (Equation~\eqref{eq:without_cfg_energy}) uses $\tau=0.55$ and $s=0.08$. The sweeps share their raw-FlowEdit endpoint. Background LPIPS is evaluated on the 556 local tasks. At $\lambda=0$ under uniform scaling, VAE encoding and decoding can still change pixels despite the absence of editing updates.

\begin{figure}[htbp]
\centering
\includegraphics[width=\linewidth]{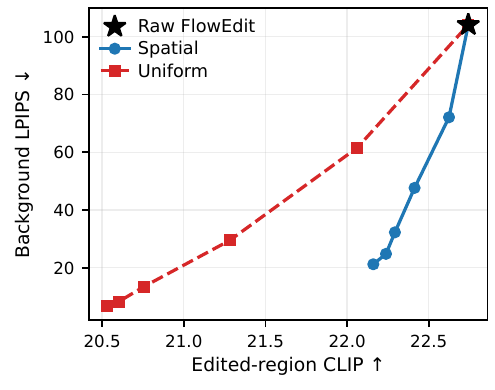}
\caption{Spatial reweighting and uniform update scaling on FLUX.1-dev. Each point is a measured control setting; the curves share the raw-FlowEdit endpoint. At comparable edited-region CLIP, spatial reweighting attains lower background LPIPS. Background LPIPS is scaled by $10^3$.}
\label{fig:edit_preservation_tradeoff}
\end{figure}

\paragraph{Results.}
At comparable edited-region CLIP, spatial reweighting achieves substantially lower background LPIPS than uniform scaling (Figure~\ref{fig:edit_preservation_tradeoff}). Thus, spatial selectivity contributes to the preservation gains in Table~\ref{tab:main_results} beyond the effect of uniformly reducing editing strength.

\subsection{Threshold: Localization and Downstream Editing}
\label{app:tau_sensitivity}

\begin{figure*}[htbp]
\centering
\includegraphics[width=\textwidth]{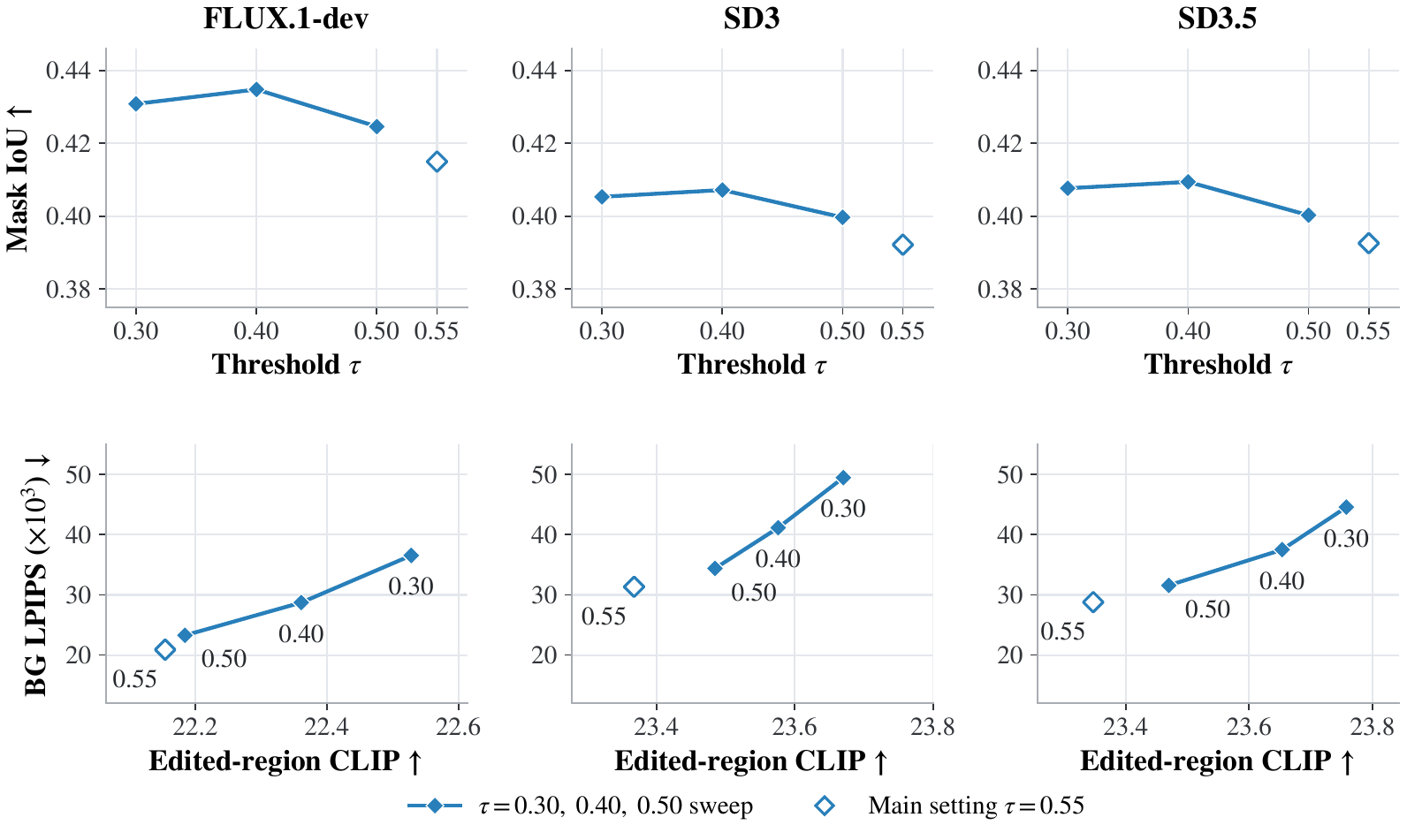}
\caption{Threshold sensitivity across three backbones at fixed $s=0.08$. Top: $\tau$ versus mask IoU. Bottom: edited-region CLIP versus background LPIPS, with points labeled by $\tau$. Both rows show $0.30/0.40/0.50/0.55$: solid lines connect the three settings evaluated with reconstructed quantized priors, while unconnected hollow markers show the $0.55$ reference from an earlier run with the original (unquantized) priors. IoU and background LPIPS use 556 local tasks; CLIP uses 700 tasks. }
\label{fig:parameter_tau_scan}
\end{figure*}

The threshold changes which locations receive large editing weights. We evaluate localization and downstream editing at $\tau\in\{0.30,0.40,0.50\}$ with fixed $s=0.08$ across FLUX.1-dev, SD3, and SD3.5. Both rows of Figure~\ref{fig:parameter_tau_scan} additionally show $\tau=0.55$ as a separate reference from an earlier run with the original (unquantized) priors.

The three-point sweep reconstructs the aggregated responses $E$ from saved quantized response/mask images, rather than recovering the original floating-point maps. The underlying priors use $E$, $5\times5$ smoothing, and 10th/90th-percentile normalization, with recorded intermediate windows at indices 7--21 for FLUX.1-dev and 17--38 for SD3/SD3.5 (zero-based). We evaluate localization by thresholding each soft prior at 0.5 and computing IoU on 556 local PIE-Bench tasks. For editing, we generate 700 outputs per setting; CLIP averages 700 tasks, and background LPIPS uses the 556 local tasks.

Among the displayed thresholds, $\tau=0.40$ yields the highest mask IoU on all three backbones, whereas $\tau=0.55$ favors more conservative spatial coverage. In the downstream sweep, increasing $\tau$ lowers both CLIP and background LPIPS: tighter spatial control improves preservation but limits the semantic change. The threshold with the highest binary-mask IoU therefore need not provide the desired final editing--preservation balance. We use $\tau=0.55$ as a preservation-oriented operating point, not as \mbox{a universal optimum}.

\subsection{Softness: Editing Changes despite an Unchanged Binary Mask}
\label{app:softness_sensitivity}

For $s>0$, the sigmoid prior satisfies
\begin{equation}
\begin{gathered}
M(x)=\sigma\!\left(\frac{\widetilde E(x)-\tau}{s}\right),\\
M(x)>0.5\ \Longleftrightarrow\ \widetilde E(x)>\tau.
\end{gathered}
\label{eq:softness_binary_invariance}
\end{equation}
Thus, changing $s$ does not change the hard mask evaluated at threshold 0.5, although it changes the weights applied during editing. An offline FLUX.1-dev check at $\tau=0.55$ with $s\in\{0.04,0.06,0.08,0.12,0.16\}$ yields the same IoU, precision, and recall (0.4150, 0.6130, and 0.6293) for every setting.

To evaluate the effect on final images, we separately fix $\tau=0.30$ and sweep $s\in\{0.04,0.08,0.16,0.32\}$ on FLUX.1-dev. We use the reconstructed priors and default CFG $(1.5,5.5)$; the $s=0.08$ outputs are reused from the threshold sweep. All plotted metrics average the same 556 local tasks. Consequently, the CLIP values are not directly comparable to the 700-task averages in Figure~\ref{fig:parameter_tau_scan}.

\begin{figure}[htbp]
\centering
\includegraphics[width=\linewidth]{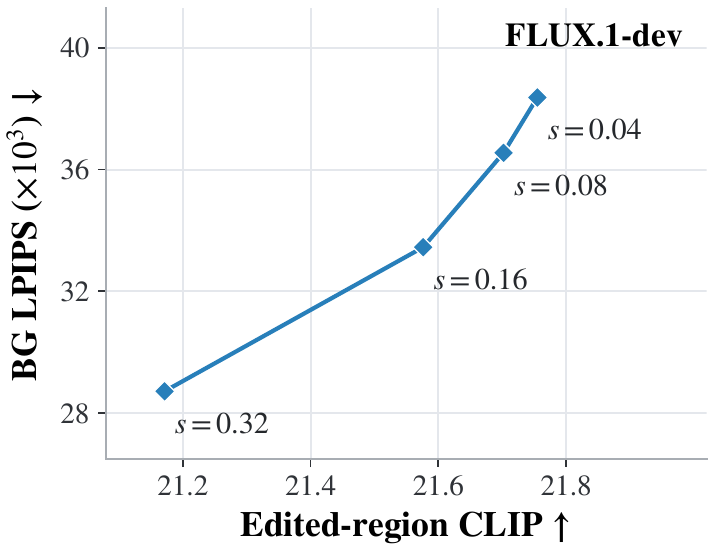}
\caption{Softness sensitivity on FLUX.1-dev at fixed $\tau=0.30$. Point labels give $s$; both edited-region CLIP and background LPIPS use the same 556 local tasks. Increasing $s$ in this sweep improves preservation while lowering CLIP, although the binary mask at threshold 0.5 is unchanged.}
\label{fig:parameter_s_scan}
\end{figure}

Figure~\ref{fig:parameter_s_scan} shows that $s$ changes the editing--preservation trade-off even when hard-mask localization is invariant. Increasing $s$ lowers both CLIP and background LPIPS in this sweep. The softness parameter must therefore be assessed through the final edits, rather than binary IoU alone.

\subsection{Aggregation-window Sensitivity}
\label{app:aggregation_window_sensitivity}

We keep the full FLUX.1-dev prior-extraction trajectory fixed (28 scheduled steps, with 24 actual updates at indices 4--27) and aggregate three subsets of its velocity differences: all available steps (4--27), an intermediate window (7--21), and a narrower intermediate window (11--17). We use $\tau=0.55$, $s=0.08$, and default CFG $(1.5,5.5)$ for prior-guided editing. This changes the responses included in the prior, not the number of updates executed during prior extraction.

\begin{table*}[htbp]
\centering
\caption{Aggregation-window sensitivity on FLUX.1-dev. Indices are zero-based; the prior-extraction trajectory executes all 24 updates for every row. IoU is evaluated on 556 local tasks using masks from the same trajectory. Edited CLIP uses 700 tasks and background LPIPS uses the 556 local tasks. $\dagger$: downstream editing reuses an earlier run with the original (unquantized) priors; its mask differs from the intermediate-window mask used for the paired IoU comparison.}
\label{tab:aggregation_window_sensitivity}
\small
\setlength{\tabcolsep}{5pt}
\begin{tabular}{lccccc}
\toprule
Window & Indices & Responses & IoU $\uparrow$ & Edited CLIP $\uparrow$ & BG LPIPS$_{\times10^3}$ $\downarrow$ \\
\midrule
Wide & 4--27 & 24 & 0.4028 & 22.0479 & 21.1075 \\
Intermediate$^\dagger$ & 7--21 & 15 & 0.4096 & 22.1543 & 20.8951 \\
Narrow & 11--17 & 7 & 0.4118 & 22.0093 & 21.2234 \\
\bottomrule
\end{tabular}
\end{table*}

The small differences in localization and editing performance across the tested windows (Table~\ref{tab:aggregation_window_sensitivity}) support the stability of the spatial localization signals in w/o~CFG velocity differences and their robustness to the aggregation-window choice.

\subsection{Ground-truth-mask Diagnostic}
\label{app:gt_mask_tradeoff}

We also constrain FLUX.1-dev updates with the binary ground-truth edit mask, resized to latent resolution without additional dilation or smoothing. This diagnostic uses CFG $(1.5,5.5)$ and a 28-step schedule with 24 actual updates.

\begin{table*}[htbp]
\centering
\caption{Ground-truth (GT) mask diagnostic on FLUX.1-dev. Constraining edits with annotated regions improves preservation while also reducing CLIP similarity. Background SSIM, LPIPS, and MSE are scaled by $10^2$, $10^3$, and $10^4$, respectively.}
\label{tab:gt_mask_tradeoff}
\resizebox{\textwidth}{!}{%
\begin{tabular}{lrrrrrrr}
\toprule
Method & Edited CLIP $\uparrow$ & Whole CLIP $\uparrow$ & BG SSIM $\uparrow$ & BG LPIPS $\downarrow$ & BG PSNR $\uparrow$ & BG MSE $\downarrow$ & Structure $\downarrow$ \\
\midrule
FlowEdit & 22.71 & 25.88 & 83.87 & 104.79 & 21.72 & 98.32 & 28.99 \\
FlowEdit + GT mask & 22.41 & 25.27 & 96.23 & 7.76 & 36.44 & 4.46 & 9.75 \\
\bottomrule
\end{tabular}%
}
\end{table*}

Even with annotated edit regions, CLIP decreases from 22.71 to 22.41 (Table~\ref{tab:gt_mask_tradeoff}). A CLIP decrease alone therefore does not establish inaccurate automatic localization. The GT mask also defines the background evaluation region, giving this variant an oracle advantage; its preservation scores serve as diagnostic references rather than a fair baseline for automatic methods.

\section{Analysis of CFG for Localization and Editing}
\label{app:cfg_backbones}
\label{app:cfg_decomposition}

This section supplements Sections~\ref{subsec:guidance_observation} and~\ref{subsec:cfg_effects} with the CFG comparison settings, additional backbone results, the prior-extraction CFG ablation, the global norm-matching control, the CFG velocity-difference derivation, and regional-norm analysis of the decomposition terms.

\subsection{Experimental Settings for CFG Comparisons}
\label{app:cfg_experimental_settings}

This subsection describes the CFG comparison settings used in Figure~\ref{fig:cfg_localization} and Appendix~\ref{app:cfg_additional_backbones}.

For the direct-editing comparisons, we evaluate FlowEdit on the same 700 PIE-Bench tasks at $512\times512$ resolution, with seed 42 and one noise sample per step. Within each backbone, the configurations use the same input images and paired noise draws. Source and target branches share noise within a step, and fresh noise is sampled across steps. We compare w/o~CFG $(1,1)$, default CFG, and three symmetric settings $g_{\mathrm{src}}=g_{\mathrm{tar}}=g$ (Table~\ref{tab:cfg_experimental_settings}). Images are generated directly without a spatial mask. The prior is subsequently extracted from each run's velocity differences to evaluate localization.

\begin{table*}[htbp]
\centering
\caption{CFG settings and sampling schedules. Guidance pairs are (source, target). All backbones also use w/o~CFG $(1,1)$. The final column distinguishes scheduled steps from actual updates.}
\label{tab:cfg_experimental_settings}
\small
\setlength{\tabcolsep}{6pt}
\begin{tabular}{lccc}
\toprule
Backbone & Default CFG & Symmetric $g$ & Scheduled / actual steps \\
\midrule
FLUX.1-dev & $(1.5,5.5)$ & $1.5,\ 3.5,\ 5.5$ & 28 / 24 \\
SD3 & $(3.5,13.5)$ & $3.5,\ 8.5,\ 13.5$ & 50 / 33 \\
SD3.5 & $(3.5,13.5)$ & $3.5,\ 8.5,\ 13.5$ & 50 / 33 \\
\bottomrule
\end{tabular}
\end{table*}

\paragraph{Prior construction and evaluation.}
We use the aggregated response $E$ (Equation~\eqref{eq:without_cfg_energy}), $5\times5$ smoothing, 10th/90th-percentile normalization, and the same sigmoid parameters $\tau=0.55$ and $s=0.08$. The recorded aggregation windows are zero-based indices 7--21 for FLUX.1-dev and 17--38 for SD3/SD3.5. We binarize each soft prior at 0.5 and compare it with the ground-truth edit mask. IoU and background metrics use the 556 local tasks; structure distance is evaluated over the full image for those same local tasks. Displayed CLIP and SSIM are scaled by $10^2$, LPIPS and structure distance by $10^3$, and MSE by $10^4$; PSNR is measured in dB.

\subsection{Additional Results on FLUX.1-dev and SD3.5}
\label{app:cfg_additional_backbones}

\begin{figure*}[htbp]
\centering
\includegraphics[width=0.85\textwidth]{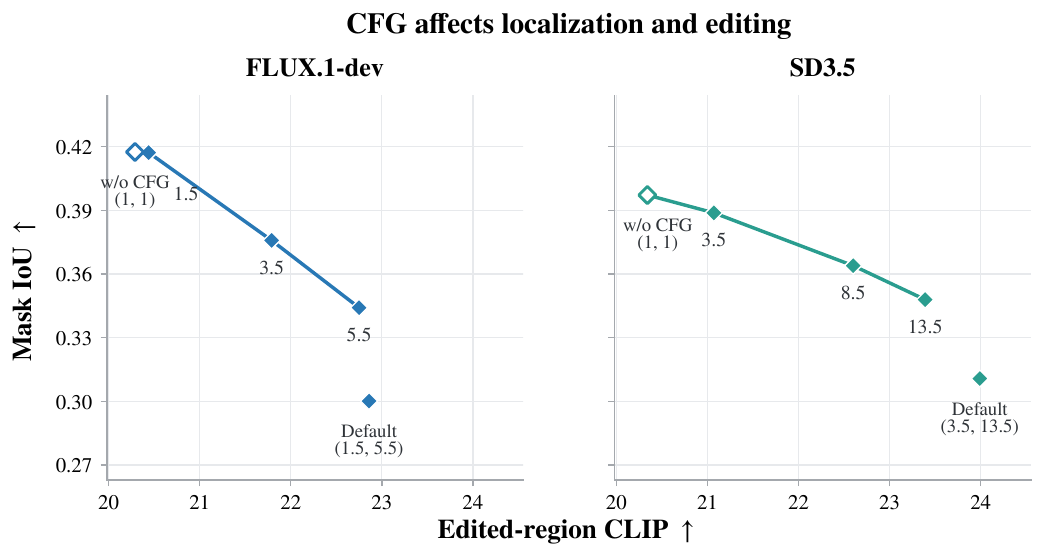}
\caption{Additional CFG comparisons on FLUX.1-dev and SD3.5. Point labels give symmetric guidance strength $g$; the w/o~CFG and default CFG points show their complete (source, target) settings. Connected points form the symmetric guidance sweep. All edited images are generated without a spatial mask; mask IoU evaluates the prior extracted from the corresponding velocity differences.}
\label{fig:cfg_backbones}
\end{figure*}

Figure~\ref{fig:cfg_backbones} extends the SD3 comparison in Section~\ref{subsec:cfg_effects}. Both backbones exhibit the same trends: w/o~CFG has the highest prior IoU but the lowest edited-region CLIP among the tested configurations, whereas default CFG has the highest CLIP and lowest IoU. Increasing symmetric guidance raises CLIP while reducing IoU.

\subsection{Effect of Prior-extraction CFG}
\label{app:prior_guidance_downstream}

We evaluate this ablation on the same 556 local-edit tasks in PIE-Bench, excluding tasks whose ground-truth masks cover the entire image. All seven metrics, including both CLIP scores, are computed on this common subset. We compare a prior-extraction stage using w/o~CFG $(1,1)$ with one using the backbone's default CFG: $(1.5,5.5)$ for FLUX.1-dev and $(3.5,13.5)$ for SD3/SD3.5. The prior-guided editing stage uses default CFG in both cases, with the same downstream sampling settings within each backbone. Both prior-extraction configurations use the full schedules and aggregation windows specified in Appendix~\ref{app:cfg_experimental_settings}, with $\tau=0.55$ and $s=0.08$. The resulting soft prior is fixed throughout the editing stage. FLUX.1-dev uses saved 8-bit soft masks, while SD3/SD3.5 use float32 priors.

\begin{table*}[htbp]
\centering
\caption{Effect of prior-extraction CFG with prior-guided editing fixed at default CFG. All metrics are evaluated on the same 556 local-edit tasks in PIE-Bench. Metric scaling follows Appendix~\ref{app:cfg_experimental_settings}.}
\label{tab:prior_guidance_downstream}
\small
\setlength{\tabcolsep}{3pt}
\begin{tabular*}{\textwidth}{@{\extracolsep{\fill}}llrrrrrrr@{}}
\toprule
\multirow{2}{*}{Backbone} & \multirow{2}{*}{Prior CFG} & \multirow{2}{*}{Structure $\downarrow$} & \multicolumn{4}{c}{Background preservation} & \multicolumn{2}{c}{CLIP $\uparrow$} \\
\cmidrule(lr){4-7}\cmidrule(lr){8-9}
& & & PSNR $\uparrow$ & LPIPS $\downarrow$ & MSE $\downarrow$ & SSIM $\uparrow$ & Whole & Edited \\
\midrule
\multirow{2}{*}{FLUX.1-dev} & w/o~CFG & 7.860 & 30.538 & 20.895 & 19.225 & 94.544 & 25.133 & 21.231 \\
 & default CFG & 12.179 & 25.968 & 41.770 & 45.080 & 91.744 & 25.256 & 21.228 \\
\midrule
\multirow{2}{*}{SD3} & w/o~CFG & 10.441 & 27.991 & 31.817 & 28.740 & 91.695 & 26.186 & 22.095 \\
 & default CFG & 15.647 & 24.828 & 54.023 & 54.934 & 89.167 & 26.367 & 22.121 \\
\midrule
\multirow{2}{*}{SD3.5} & w/o~CFG & 9.163 & 28.783 & 29.214 & 24.850 & 92.030 & 26.319 & 22.106 \\
 & default CFG & 13.389 & 25.673 & 48.216 & 45.470 & 89.836 & 26.586 & 22.129 \\
\bottomrule
\end{tabular*}
\end{table*}

As shown in Table~\ref{tab:prior_guidance_downstream}, extracting priors under w/o~CFG reduces background LPIPS by 50.0\%, 41.1\%, and 39.4\% on FLUX.1-dev, SD3, and SD3.5, respectively, and improves all other preservation metrics. The corresponding relative changes in edited-region CLIP are $+0.013\%$, $-0.121\%$, and $-0.103\%$. All relative changes use the corresponding default-CFG-prior result as the reference. These results support using w/o~CFG for prior extraction while retaining default CFG for semantic editing.

\subsection{Per-step Global Norm Matching}
\label{app:global_norm_matching}

We test whether a single global rescaling of the w/o~CFG update can recover default CFG editing. At every step of the norm-matched trajectory, we evaluate the w/o~CFG update $\mathbf U_t$ and default CFG update $\mathbf H_t$ at the \emph{same current source/target states and with the same sampled noise}. We advance the state using only
\begin{equation}
\widehat{\mathbf U}_t
=\frac{\|\mathbf H_t\|_{\mathrm F}}
{\|\mathbf U_t\|_{\mathrm F}+10^{-8}}\,\mathbf U_t.
\label{eq:cfg_global_norm_matching}
\end{equation}
The Frobenius norm covers all spatial positions and channels, yielding one scalar for the entire update. Both updates are recomputed at the next state; the reference magnitude is not taken from a separately generated default-CFG trajectory. We use the same sampling settings as in Table~\ref{tab:cfg_experimental_settings}.

\begin{table*}[htbp]
\centering
\caption{Direct editing with w/o~CFG, global norm matching, and default CFG. Both CLIP metrics use the actual outputs of all 700 tasks; background and structure metrics use the 556 local tasks. The scale factors are those specified in Appendix~\ref{app:cfg_experimental_settings}.}
\label{tab:cfg_norm_editing}
\small
\setlength{\tabcolsep}{3pt}
\begin{tabular*}{\textwidth}{@{\extracolsep{\fill}}llrrrrrrr@{}}
\toprule
\multirow{2}{*}{Backbone} & \multirow{2}{*}{Update} & \multirow{2}{*}{Structure $\downarrow$} & \multicolumn{4}{c}{Background preservation} & \multicolumn{2}{c}{CLIP $\uparrow$} \\
\cmidrule(lr){4-7}\cmidrule(lr){8-9}
& & & PSNR $\uparrow$ & LPIPS $\downarrow$ & MSE $\downarrow$ & SSIM $\uparrow$ & Whole & Edited \\
\midrule
\multirow{3}{*}{FLUX.1-dev} & w/o~CFG & 0.95 & 36.30 & 7.77 & 4.70 & 96.32 & 23.47 & 20.29 \\
 & Norm-matched & 42.84 & 20.34 & 95.92 & 142.43 & 85.35 & 25.13 & 21.30 \\
 & default CFG & 28.99 & 21.72 & 104.79 & 98.32 & 83.87 & 25.88 & 22.71 \\
\midrule
\multirow{3}{*}{SD3} & w/o~CFG & 0.97 & 35.32 & 10.23 & 6.11 & 94.31 & 23.49 & 20.33 \\
 & Norm-matched & 84.07 & 16.76 & 178.42 & 298.79 & 75.39 & 24.50 & 20.67 \\
 & default CFG & 26.39 & 22.27 & 100.15 & 85.71 & 83.93 & 27.51 & 23.92 \\
\midrule
\multirow{3}{*}{SD3.5} & w/o~CFG & 0.96 & 35.26 & 10.48 & 6.09 & 94.22 & 23.49 & 20.34 \\
 & Norm-matched & 64.17 & 18.39 & 149.86 & 214.27 & 78.70 & 24.60 & 20.80 \\
 & default CFG & 22.55 & 23.27 & 88.78 & 69.66 & 85.57 & 27.62 & 23.95 \\
\bottomrule
\end{tabular*}
\end{table*}

Global norm matching increases edited-region CLIP relative to w/o~CFG, but yields lower CLIP and higher structure distance than default CFG on all three backbones (Table~\ref{tab:cfg_norm_editing}). SD3 and SD3.5 also deteriorate in all four background metrics; FLUX.1-dev has mixed background results. The norm-matched edits exhibit the severe blur shown in Figure~\ref{fig:decouple_qualitative}. Matching total update magnitude is therefore insufficient to recover the editing quality of default CFG.


\subsection{Derivation of the CFG Velocity-Difference Decomposition}
\label{app:cfg_decomposition_derivation}

We provide the derivation of the decomposition in
Section~\ref{subsec:guidance_observation} under standard linear CFG.
Fix a time $t$ and the paired source and target states
$\mathbf z_t^{\mathrm{src}}$ and $\mathbf z_t^{\mathrm{tar}}$,
with corresponding text prompts $P_{\mathrm{src}}$ and $P_{\mathrm{tar}}$.
Each branch is evaluated at its own state, and this pair of states is held
fixed throughout the derivation.

For a latent state $\mathbf z$ and a text prompt $P$, standard linear CFG
with scale $g$ gives
\begin{equation}
\begin{aligned}
\mathbf V_\theta^g(\mathbf z,t\mid P)
&=\mathbf V_\theta(\mathbf z,t\mid\emptyset)\\
&\quad+g\bigl[
\mathbf V_\theta(\mathbf z,t\mid P)
-\mathbf V_\theta(\mathbf z,t\mid\emptyset)
\bigr]\\
&=\mathbf V_\theta(\mathbf z,t\mid P)\\
&\quad+(g-1)\bigl[
\mathbf V_\theta(\mathbf z,t\mid P)\\
&\hphantom{{}\quad+(g-1)\bigl[}
-\mathbf V_\theta(\mathbf z,t\mid\emptyset)
\bigr],
\end{aligned}
\label{eq:app_cfg_linear_guidance}
\end{equation}
where $\emptyset$ denotes the empty text condition used for the unconditional
prediction. In Section~\ref{subsec:guidance_observation}, we suppress the state
and time arguments and abbreviate the conditional and unconditional predictions
as $\mathbf V_\theta^P$ and $\mathbf V_\theta^u$, respectively.

The source and target guidance directions are therefore
\begin{equation}
\begin{aligned}
\mathbf d_t^{\mathrm{src}}
&=\mathbf V_\theta(\mathbf z_t^{\mathrm{src}},t\mid P_{\mathrm{src}})\\
&\quad-\mathbf V_\theta(\mathbf z_t^{\mathrm{src}},t\mid\emptyset),\\
\mathbf d_t^{\mathrm{tar}}
&=\mathbf V_\theta(\mathbf z_t^{\mathrm{tar}},t\mid P_{\mathrm{tar}})\\
&\quad-\mathbf V_\theta(\mathbf z_t^{\mathrm{tar}},t\mid\emptyset).
\end{aligned}
\label{eq:app_cfg_guidance_directions}
\end{equation}
The unconditional predictions are evaluated at the respective branch states
and are not assumed to be identical.

For guidance scales $\mathbf g=(g_{\mathrm{src}},g_{\mathrm{tar}})$,
subtracting the guided source velocity from the guided target velocity yields
\begin{equation}
\begin{aligned}
\Delta\mathbf V_t^{\mathbf g}
&=\mathbf V_\theta^{g_{\mathrm{tar}}}
(\mathbf z_t^{\mathrm{tar}},t\mid P_{\mathrm{tar}})\\
&\quad-\mathbf V_\theta^{g_{\mathrm{src}}}
(\mathbf z_t^{\mathrm{src}},t\mid P_{\mathrm{src}})\\
&=\Delta\mathbf V_t^{\mathrm{w/oCFG}}
+(g_{\mathrm{tar}}-1)\mathbf d_t^{\mathrm{tar}}\\
&\quad-(g_{\mathrm{src}}-1)\mathbf d_t^{\mathrm{src}},
\end{aligned}
\label{eq:app_cfg_velocity_difference}
\end{equation}
where $\Delta\mathbf V_t^{\mathrm{w/oCFG}}$ denotes the target conditional velocity
minus the source conditional velocity at the same paired states, corresponding
to $g_{\mathrm{src}}=g_{\mathrm{tar}}=1$.

Adding and subtracting $(g_{\mathrm{tar}}-1)\mathbf d_t^{\mathrm{src}}$,
we obtain
\begin{equation}
\begin{aligned}
&(g_{\mathrm{tar}}-1)\mathbf d_t^{\mathrm{tar}}
-(g_{\mathrm{src}}-1)\mathbf d_t^{\mathrm{src}}\\
&\quad=
(g_{\mathrm{tar}}-1)
\left(\mathbf d_t^{\mathrm{tar}}-\mathbf d_t^{\mathrm{src}}\right)\\
&\qquad+\left[(g_{\mathrm{tar}}-1)-(g_{\mathrm{src}}-1)\right]
\mathbf d_t^{\mathrm{src}}\\
&\quad=
(g_{\mathrm{tar}}-1)
\left(\mathbf d_t^{\mathrm{tar}}-\mathbf d_t^{\mathrm{src}}\right)\\
&\qquad+(g_{\mathrm{tar}}-g_{\mathrm{src}})
\mathbf d_t^{\mathrm{src}}.
\end{aligned}
\label{eq:app_cfg_regrouping}
\end{equation}
Substitution gives the decomposition used in the main text:
\begin{equation}
\begin{aligned}
\Delta\mathbf V_t^{\mathbf g}
&=\Delta\mathbf V_t^{\mathrm{w/oCFG}}\\
&\quad+\underbrace{(g_{\mathrm{tar}}-g_{\mathrm{src}})\mathbf d_t^{\mathrm{src}}}_{\text{asymmetry term}}\\
&\quad+\underbrace{(g_{\mathrm{tar}}-1)\left(\mathbf d_t^{\mathrm{tar}}-\mathbf d_t^{\mathrm{src}}\right)}_{\text{amplification term}}.
\end{aligned}
\label{eq:app_cfg_decomposition}
\end{equation}

For symmetric guidance, $g_{\mathrm{src}}=g_{\mathrm{tar}}=g$,
the asymmetry term vanishes and the amplification term becomes
$(g-1)(\mathbf d_t^{\mathrm{tar}}-\mathbf d_t^{\mathrm{src}})$.
At $(1,1)$, both additional terms vanish.

This identity compares guidance configurations at a fixed pair of states.
Trajectories evolved under different guidance configurations can visit
different states; the w/o~CFG term in the decomposition is evaluated at the
current paired states rather than taken from a separately evolved trajectory.

\subsection{Empirical Analysis of the CFG Decomposition}
\label{app:cfg_decomposition_empirical}

We empirically examine the decomposition in Equation~\eqref{eq:cfg_decomposition}, repeated here for convenience:
\begin{equation}
\resizebox{\dimexpr\linewidth-2.5em\relax}{!}{$\displaystyle
\Delta\mathbf V_t^{\mathbf g}
=\underbrace{\Delta\mathbf V_t^{\mathrm{w/oCFG}}}_{\mathbf U_t}
+\underbrace{(g_{\mathrm{tar}}-g_{\mathrm{src}})\mathbf d_t^{\mathrm{src}}}_{\mathbf A_t}
+\underbrace{(g_{\mathrm{tar}}-1)\left(\mathbf d_t^{\mathrm{tar}}-\mathbf d_t^{\mathrm{src}}\right)}_{\mathbf B_t}.
$}
\label{eq:app_cfg_empirical_terms}
\end{equation}
We refer to $\mathbf U_t$, $\mathbf A_t$, and $\mathbf B_t$ as the base, asymmetry, and amplification terms, respectively. The guidance directions $\mathbf d_t^{\mathrm{src}}$ and $\mathbf d_t^{\mathrm{tar}}$ are the conditional-minus-unconditional velocities evaluated at their respective branch states. All three terms are evaluated at the current paired states of the trajectory being analyzed. Thus, the base term on a guided trajectory need not equal the velocity difference on a separately evolved w/o~CFG trajectory.

\paragraph{Experimental setup and regional statistics.}
We evaluate SD3 on the same 556 PIE-Bench local-edit tasks using the original prompts, a resolution of $512\times512$, and seed 42. We consider five CFG configurations: $(1,1)$, $(3.5,3.5)$, $(8.5,8.5)$, $(13.5,13.5)$, and the default configuration $(3.5,13.5)$. Each configuration follows its own FlowEdit trajectory. The 50-step schedule executes updates at indices 17--49, and we report statistics over indices 17--38, inclusive.

Ground-truth masks mapped to the latent grid define the edit region $\Omega_E$ and its complement $\Omega_N$. For each term, we first compute the channel-wise L2 norm at each spatial location, average within each region, and then average equally over the selected steps and samples. We denote the resulting edit- and non-edit-region means by $\mu_E$ and $\mu_N$, and report their ratio:
\begin{equation}
\rho=\frac{\mu_N}{\mu_E}.
\label{eq:app_cfg_regional_ratio}
\end{equation}
A lower ratio indicates stronger edit/non-edit response contrast. Total denotes the full velocity difference $\Delta\mathbf V_t^{\mathbf g}$, whose norm is computed after summing the three vector fields. Table~\ref{tab:cfg_decomposition_norms} reports the results.

\begin{table*}[t]
\centering
\caption{Regional norms of the CFG decomposition terms along five SD3 FlowEdit trajectories. For each term, we report the edit-region mean $\mu_E$, non-edit-region mean $\mu_N$, and their ratio $\rho=\mu_N/\mu_E$. Ratios are computed from unrounded means. ``--'' denotes the undefined ratio of a zero term.}
\label{tab:cfg_decomposition_norms}
\small
\setlength{\tabcolsep}{3pt}
\begin{tabular*}{\textwidth}{@{\extracolsep{\fill}}lrrrrrrrrrrrr@{}}
\toprule
\multirow{2}{*}{CFG $(g_{\mathrm{src}},g_{\mathrm{tar}})$}
& \multicolumn{3}{c}{Base $\mathbf U$}
& \multicolumn{3}{c}{Asymmetry $\mathbf A$}
& \multicolumn{3}{c}{Amplification $\mathbf B$}
& \multicolumn{3}{c}{Total} \\
\cmidrule(lr){2-4}\cmidrule(lr){5-7}\cmidrule(lr){8-10}\cmidrule(lr){11-13}
& $\mu_E$ & $\mu_N$ & $\rho$
& $\mu_E$ & $\mu_N$ & $\rho$
& $\mu_E$ & $\mu_N$ & $\rho$
& $\mu_E$ & $\mu_N$ & $\rho$ \\
\midrule
$(1,1)$ & 0.327 & 0.123 & 0.3764 & 0.000 & 0.000 & -- & 0.000 & 0.000 & -- & \textbf{0.327} & \textbf{0.123} & \textbf{0.3764} \\
$(3.5,3.5)$ & 0.529 & 0.201 & 0.3796 & 0.000 & 0.000 & -- & 0.787 & 0.342 & 0.4337 & 1.210 & 0.499 & 0.4122 \\
$(8.5,8.5)$ & 0.922 & 0.414 & 0.4489 & 0.000 & 0.000 & -- & 2.814 & 1.421 & 0.5049 & 3.340 & 1.645 & 0.4924 \\
$(13.5,13.5)$ & 1.270 & 0.658 & 0.5179 & 0.000 & 0.000 & -- & 5.153 & 2.865 & 0.5560 & 5.725 & 3.148 & 0.5498 \\
$(3.5,13.5)$ (default) & 1.183 & 0.721 & 0.6091 & 3.650 & 2.334 & 0.6394 & 5.362 & 3.174 & 0.5920 & 4.459 & 2.926 & 0.6562 \\

\bottomrule
\end{tabular*}
\end{table*}

\paragraph{Base response under w/o CFG.}
Under w/o~CFG, both the asymmetry and amplification terms vanish, so the total update consists only of the base term. Its mean magnitude is 0.327 in the edit region and 0.123 in the non-edit region, giving a non-edit-to-edit ratio of 0.3764. The response is therefore substantially stronger inside the intended edit region, providing a spatial signal for distinguishing edit from non-edit regions.

\paragraph{Amplification under symmetric CFG.}
Under symmetric CFG, the asymmetry term remains zero, while the amplification term becomes nonzero. At all three tested symmetric scales, the amplification term has a higher non-edit-to-edit ratio than the base term, indicating weaker regional contrast. As the common scale increases from 3.5 to 13.5, its magnitude relative to the base term rises from 1.49 to 4.06 in the edit region and from 1.70 to 4.36 in the non-edit region.

The base term also changes along these trajectories. Although it contains no explicit guidance coefficient, earlier updates alter the states at which it is subsequently evaluated. Its regional means are identical across configurations at the first executed step and diverge from the next step onward. Across the three symmetric settings, its ratio increases from 0.3796 to 0.4489 and 0.5179. Thus, stronger symmetric guidance introduces an increasingly large amplification term with weaker regional contrast and also changes the subsequent base-term responses through the trajectory. The total response ratio correspondingly increases from 0.4122 to 0.4924 and 0.5498.

\paragraph{Asymmetry under default CFG.}
Default CFG additionally retains the asymmetry term. Its mean magnitudes are 3.650 in the edit region and 2.334 in the non-edit region, approximately three times those of the base term. Its non-edit-to-edit ratio is also higher than that of the base term (0.6394 vs.\ 0.6091), indicating a large additional response with weaker regional contrast.

The base term along the default CFG trajectory has a ratio of 0.6091, compared with 0.3764 along the w/o~CFG trajectory, further illustrating the influence of guidance history on subsequent base-term responses. After the three vector components combine, the total ratio reaches 0.6562, the highest among the tested configurations. This ratio exceeds those of the individual terms, showing that cancellation between components does not necessarily improve the regional contrast of the final update.

\paragraph{Implications for localization.}
Among the tested configurations, w/o~CFG yields the lowest non-edit-to-edit ratio of the total response and therefore the strongest regional response contrast. This agrees with its highest mask IoU in Figure~\ref{fig:cfg_localization}. Both additional terms remain zero throughout the w/o~CFG trajectory, avoiding their direct contributions and their accumulated influence on subsequent states. These results support using an independent w/o~CFG trajectory to extract the spatial localization prior, which then constrains the effective semantic updates under default CFG.

\section{Multi-seed Main Results}
\label{app:multi_seed}

To assess the stability of the quantitative results, we repeat the main experiments five times with different random seeds. The random seeds affect the stochastic components of the editing process, including the noise samples used in FlowEdit-style updates and prior extraction. We report the mean and the run-to-run standard deviation over these five runs. The repeated experiments follow the same evaluation protocol as the main paper, using PIE-Bench and the same metrics for structure distance, background preservation, and semantic editing fidelity.

\begin{table*}[t]
\centering
\caption{Multi-seed quantitative results on PIE-Bench. We report mean $\pm$ standard deviation over five runs with different random seeds. Lower is better for Structure Distance, LPIPS, and MSE; higher is better for PSNR, SSIM, and CLIP similarity. }
\label{tab:multi_seed_results}
\resizebox{\textwidth}{!}{
\begin{tabular}{llccccccc}
\toprule
\multirow{2}{*}{\textbf{Method}}
& \multirow{2}{*}{\textbf{Backbone}}
& \textbf{Structure}
& \multicolumn{4}{c}{\textbf{Background Preservation}}
& \multicolumn{2}{c}{\textbf{CLIP Similarity}} \\
\cmidrule(lr){3-3}
\cmidrule(lr){4-7}
\cmidrule(lr){8-9}
& & Distance $\downarrow$
& PSNR $\uparrow$
& LPIPS$_{\times 10^3}$ $\downarrow$
& MSE$_{\times 10^4}$ $\downarrow$
& SSIM$_{\times 100}$ $\uparrow$
& Whole $\uparrow$
& Edited $\uparrow$ \\
\midrule
FlowEdit & FLUX.1-dev
& $28.99 \pm 0.24$
& $21.72 \pm 0.06$
& $104.79 \pm 0.06$
& $98.32 \pm 0.26$
& $83.87 \pm 0.02$
& $25.88 \pm 0.01$
& $22.71 \pm 0.04$ \\
Ours & FLUX.1-dev
& $7.86 \pm 0.21$
& $30.54 \pm 0.05$
& $20.90 \pm 0.05$
& $19.23 \pm 0.22$
& $94.54 \pm 0.02$
& $25.25 \pm 0.01$
& $22.15 \pm 0.03$ \\
\midrule
FlowEdit & SD3
& $26.39 \pm 0.26$
& $22.27 \pm 0.07$
& $100.15 \pm 0.07$
& $85.71 \pm 0.29$
& $83.93 \pm 0.03$
& $27.51 \pm 0.02$
& $23.92 \pm 0.04$ \\
Ours & SD3
& $10.01 \pm 0.22$
& $28.16 \pm 0.05$
& $31.34 \pm 0.05$
& $27.85 \pm 0.22$
& $91.75 \pm 0.02$
& $26.71 \pm 0.01$
& $23.37 \pm 0.03$ \\
\midrule
FlowEdit & SD3.5
& $22.55 \pm 0.25$
& $23.27 \pm 0.07$
& $88.78 \pm 0.06$
& $69.66 \pm 0.27$
& $85.57 \pm 0.02$
& $27.62 \pm 0.02$
& $23.95 \pm 0.04$ \\
Ours & SD3.5
& $8.90 \pm 0.20$
& $28.78 \pm 0.05$
& $28.79 \pm 0.05$
& $24.59 \pm 0.21$
& $92.03 \pm 0.02$
& $26.86 \pm 0.01$
& $23.35 \pm 0.03$ \\
\bottomrule
\end{tabular}}
\end{table*}

Across seeds, our method consistently improves background and structure preservation over the corresponding FlowEdit baseline. The standard deviations are small relative to the performance gains, indicating that the improvement is not caused by a particular random seed. The semantic editing metrics remain competitive, suggesting that the spatial reweighting improves preservation without substantially weakening the intended edit.

\section{Limitations}
\label{app:limitations}

\begin{figure*}[t]
    \centering
    \includegraphics[width=1.00\linewidth]{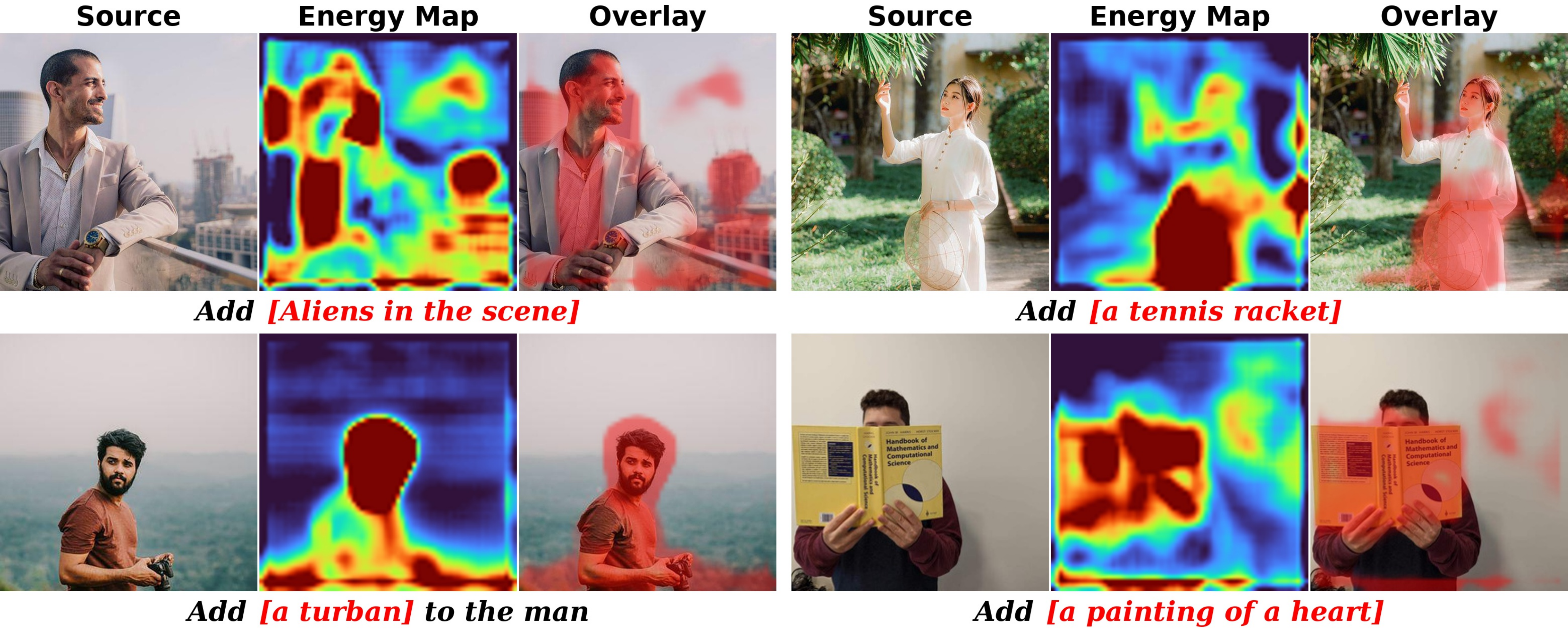}
    \caption{Limitations of spatial-prior extraction in object-add edits using FLUX.1-dev.
    }
    \label{fig:failure_1}
\end{figure*}

\begin{figure*}[t]
    \centering
    \includegraphics[width=1.00\linewidth]{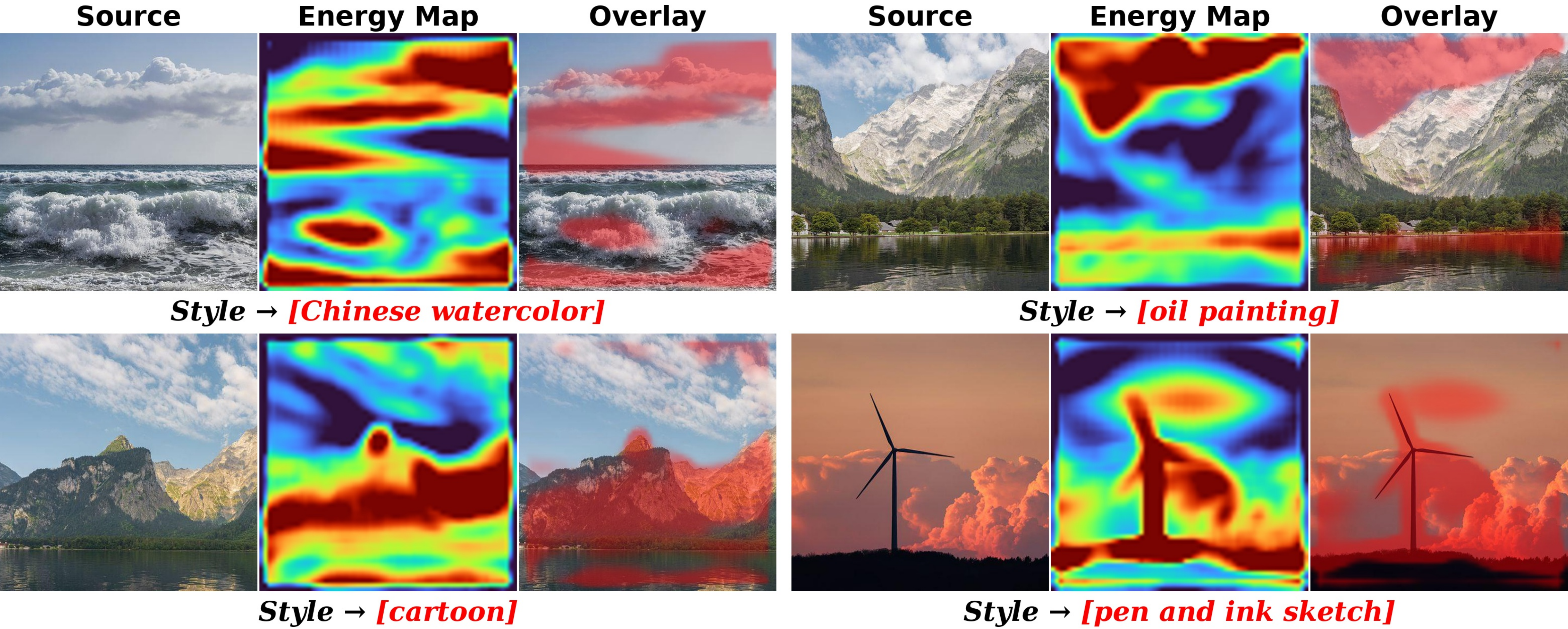}
    \caption{Limitations of spatial-prior extraction in style-change edits using FLUX.1-dev.
    }
    \label{fig:failure_2}
\end{figure*}

Our method is most reliable for local edits where the source image already contains a clear editing target or spatial anchor. In such cases, temporally aggregating FlowEdit velocity differences \(\Delta \mathbf{V}_t\) can accumulate edit-driven responses around the relevant region and produce a selective spatial prior. This reliance on a spatial anchor also reveals several limitations.

First, for \emph{object-add} edits, the target object usually has no clear source-side counterpart or spatial anchor in the source image. As a result, the edit-driven updates lack a stable region on which to accumulate across timesteps. The extracted prior may therefore become diffuse or attach to semantically related existing regions, rather than accurately localizing where the newly added object should appear, as illustrated in Figure~\ref{fig:failure_1}.

Second, for \emph{style-change} edits, the target modification is often a global or large-range appearance transformation, such as changing texture, color, brushstroke, or overall rendering style. Such edits do not necessarily require a localized mask, because the desired change is expected to affect the whole image relatively consistently. Accordingly, our temporal aggregation mechanism is less suitable for this setting: the resulting prior does not uniformly cover the image, but is instead influenced by semantic structures, local textures, and high-response regions, leading to a content-dependent and non-uniform distribution, as shown in Figure~\ref{fig:failure_2}. Therefore, these cases should not be interpreted simply as localization failures: our prior is deliberately designed for local, object-related, or spatially anchored edits, and style changes that require globally consistent modification are outside the scope of this method. For such edits, we recommend applying raw FlowEdit, whose spatially global velocity differences naturally support global transformations.

\section{Broader Impact}
\label{app:broader_impact}

This work improves training-free text-guided image editing by reducing unintended changes in background and non-edit regions. This can benefit creative editing, visual content production, and user-controlled image manipulation, especially where preserving source details is important.

However, more reliable image editing tools may also be misused to create misleading, manipulated, or deceptive visual content. The proposed method does not introduce a new generative model, but it can improve the controllability of existing pretrained image generation backbones. We encourage responsible use of the method and adherence to the usage policies and safety guidelines of the underlying pretrained generative models.

\section{Assets and Licenses}
\label{app:assets}

We use publicly available datasets, pretrained models, and baseline methods. The main benchmark is PIE-Bench, which provides source images, source and target prompts, editing instructions, and ground-truth edit masks for text-guided image editing evaluation. We use publicly available pretrained flow backbones, including FLUX.1-dev, SD3, and SD3.5, following their respective licenses and terms of use. We also compare with publicly described training-free editing baselines, including FlowEdit and recent flow-based editing methods.

All existing assets used in this paper are cited in the main text or references. The proposed method does not introduce a new dataset or a new pretrained generative model. The released asset is the anonymized implementation of our method, with instructions for reproducing \mbox{the main experiments}.


\end{document}